\documentclass[a4paper,fleqn]{cas-sc}

\usepackage{enumitem} 
\usepackage{float}
\usepackage{placeins}
\usepackage{graphicx}
\usepackage[percent]{overpic}
\usepackage{xcolor}

\begin{document}

\title[mode=title]{A Perspective on Open Challenges in Deformable Object Manipulation}

\tnotetext[1]{This research was supported by a Departmental Scholarship as part of the EPSRC DigiCORTEX Project.}

\author[1]{Ryan Paul McKenna}
\cormark[1]
\orcidauthor{0009-0000-6339-9010}{Ryan Paul McKenna}
\ead{ryan.mckenna@york.ac.uk}
\credit{Conceptualization, Methodology, Investigation, Writing – original draft}

\cortext[1]{Corresponding author}

\author[1]{John Oyekan}
\orcidauthor{0000-0001-6578-9928}{John Oyekan}
\credit{Supervision, Writing – review & editing}

\affiliation[1]{organization={Department of Computer Science, University of York},
                addressline={Heslington},
                city={York},
                postcode={YO10 5DD},
                country={United Kingdom}}

\shorttitle{Open Challenges in Deformable Object Manipulation}
\shortauthors{McKenna et~al.}

\begin{abstract}
Deformable object manipulation (DOM) represents a critical challenge in robotics, with applications spanning healthcare, manufacturing, food processing, and beyond. Unlike rigid objects, deformable objects exhibit infinite dimensionality, dynamic shape changes, and complex interactions with their environment, posing significant hurdles for perception, modeling, and control. This paper reviews the state of the art in DOM, focusing on key challenges such as occlusion handling, task generalization, and scalable, real-time solutions. It highlights advancements in multi-modal perception systems, including the integration of multi-camera setups, active vision, and tactile sensing, which collectively address occlusion and improve adaptability in unstructured environments. Cutting-edge developments in physically informed reinforcement learning (RL) and differentiable simulations are explored, showcasing their impact on efficiency, precision, and scalability. The review also emphasizes the potential of simulated expert demonstrations and generative neural networks to standardize task specifications and bridge the simulation-to-reality gap. Finally, future directions are proposed, including the adoption of graph neural networks for high-level decision-making and the creation of comprehensive datasets to enhance DOM’s real-world applicability. By addressing these challenges, DOM research can pave the way for versatile robotic systems capable of handling diverse and dynamic tasks with deformable objects.
\end{abstract}


\maketitle

\section{Introduction}
The ability to manipulate objects is a critical capability for robots, as their core function involves interacting with and influencing their environment. Over the years, significant advances have been made in robotic manipulation, particularly focusing on cost-efficient robotic arms and end-effectors designed to directly engage with the physical world \cite{kroemer2021review}. Traditionally, research has assumed that objects being manipulated are rigid \cite{zhu2022challenges}, but this assumption does not apply universally. Many objects can deform when external forces are applied. In many cases these deformations cannot be ignored during manipulation tasks.

Equipping robots to handle deformable objects (DOs) opens up numerous possibilities in various fields, including industrial manufacturing, medical surgery, food processing, and elder care. These applications promise significant economic benefits, such as reducing physical strain on workers \cite{gao2022hierarchical} \cite{li2018vision} \cite{jin2021trajectory} \cite{zhou2020practical} in manufacturing, assisting in surgical procedures, cutting labor costs in food processing \cite{shi2023robocook} \cite{xu2023roboninja}, and supporting the elderly and disabled in daily tasks \cite{narasimhan2022self} \cite{zheng2022}

Despite its importance, the manipulation of deformable objects (DOM) has been less explored compared to rigid object manipulation, largely due to the complexity of perceiving, modeling, and controlling deformable materials. In contrast to rigid objects, DOs present infinite dimensionality and intricate dynamics, which complicates robotic planning and control strategies. However, recent advances in computer graphics and machine learning offer promising tools to tackle the challenges associated with DOM. Data-driven approaches offer solutions to several limitations found in traditional methods for manipulating deformable objects (DOM) \cite{billard2019trends}, \cite{yin2021modeling}. A review by Kadi and Terzic \cite{kadi2023data} specifically examined the use of data-driven techniques for clothing manipulation. Other past reviews \cite{lv2020review}, \cite{hou2019review}, \cite{arriola2020modeling} concentrated on modeling deformable objects but did not take perception and manipulation into account. While previous reviews \cite{zhu2022challenges}, \cite{sanchez2018robotic} provided broad coverage of the subject, their emphasis was primarily on analytical methods. In contrast, the review by Yin et al. \cite{yin2021modeling} offers a more comprehensive perspective by including both analytical and data-driven approaches. This literature review focuses on addressing key challenges in deformable object manipulation (DOM), particularly those that have received less attention in existing research. One of the primary challenges in DOM is the need for generalization across different types of deformable objects \cite{gu_survey}. Current approaches are often highly specific, tailored to both the object being manipulated and the task at hand, such as assembly \cite{luo2018deep} or cloth folding \cite{sun2015accurate}. This narrow focus limits the flexibility of robotic systems, making them less adaptable to new objects or tasks without significant reprogramming. In addition to generalization, another critical challenge is bridging the gap between learning from state information and high-dimensional observations, which is essential for real-world applications \cite{lin2023softgym}. State information refers to internal simulation data, such as object position or velocity, which is typically accessible in simulated environments but difficult to obtain in real-world scenarios. In contrast, high-dimensional observations, such as images or depth maps, offer more realistic sensory input but are far more complex to process. Overcoming this gap is crucial for enabling robots to manipulate deformable objects in unstructured, real-world environments. A third challenge concerns the need for physical accuracy in DOM \cite{mataslearning}, especially in high-stakes applications like robotic surgery \cite{liang2024real}. Deformable objects exhibit highly complex and non-linear behavior, with a high number of degrees of freedom and dynamic changes in form. This complexity makes accurate physical modeling and control difficult, but precision is essential in tasks where even slight inaccuracies can lead to failure or harm. Scalability and computational efficiency also present significant hurdles \cite{arriola2020modeling}. The algorithms required to model and learn how to manipulate deformable objects are computationally expensive, slowing down research progress and innovation in this field \cite{valencia2019toward}. Additionally, fundamental issues in reinforcement learning, such as reward shaping and defining goal states, become especially prominent in the context of DOM. These issues are compounded by the need to handle the unpredictable nature of deformable objects and the variability of tasks. Therefore, this review will address the challenge of task specification, specifically how to technically define when a task is completed. For example, what does it mean for a cloth to be "folded," water to be "poured," or fruit to be "picked"? Defining these tasks in a generalizable way is an open problem at the heart of DOM research \cite{zhu2022challenges}. Without clear and adaptable task specifications, robotic systems may struggle to generalize across different deformable objects and tasks, making it difficult to scale solutions across varied applications. \\

\textbf{Key Challenges:} \label{key-challenges}

\begin{itemize}
  \item The need for Generalization.
  \item The need to close the gap between learning from state information compared to high-dimensional observation.
  \item The need for physical accuracy in manipulation.
  \item The need for efficient learning.
  \item The need for task specifications.
  \item The need for Scalability across object and task types.
\end{itemize}

We begin with an overview of core concepts and approaches to DOM and then a thorough literature review which addresses each of the aforementioned challenges.

\section{Perception}

Robots need fast, precise, and multi-modal perception capabilities to understand their environment, which is essential for performing complex manipulation tasks \cite{lee2020}. Perception's main goal is to estimate the state of an object $X$, by solving an optimization problem (Equation \ref{eq:1}) based on the observation $O$ and object representation $R$:

\begin{equation} \label{eq:1}
    X^* = \arg \min_X \| O - R(X) \|
\end{equation}

where $X$ represents the state of the object. However, accurately and efficiently representing a deformable object (DO) remains a significant challenge, and solutions are often tailored to specific applications. A common approach to represent DOs is to use particles \cite{battaglia2016}. In the perception part of this review, DOs are represented as particles unless stated otherwise. Because DOs have an infinite-dimensional state space, perceiving their deformations is especially challenging. In addition, occlusion and noise are frequent in unstructured settings, demanding robust state estimation for DOs. This section reviews visual and tactile perception. 

\subsection{Visual Perception}

A vision-based approach to estimating the state of a deformable object (DO) involves three primary stages: segmentation, detection, and tracking.

\subsubsection{Segmentation} 

Segmentation focuses on distinguishing the DO from the background in an image. Detection then estimates the state of the DO in a single image. Finally, tracking maintains an estimate of the state of the DO in multiple frames. 

Nadon et al proposes a comprehensive segmentation approach tailored for deformable objects \cite{nadon2020}. This method primarily relies on multi-view, multi-modal sensory data, integrating RGB-D sensors positioned at various angles to build a full 3D model of the objects within the scene. Using multiple viewpoints, the framework can capture a more accurate representation, addressing occlusions and perspective limitations commonly faced with single-view systems. This segmentation method supports the accurate modeling and manipulation of complex deformable objects, which can change shape dynamically and marks a key benefit for robotic tasks in unstructured environments.

While effective, the multi-view setup demands substantial processing power and precise sensor calibration, which can complicate its application in real-time, posing issues for real-world scenarios. This multisensor fusion also requires sophisticated data integration algorithms to seamlessly merge information across perspectives. Another challenge is that current object recognition datasets lack explicit deformability annotations, which can hinder segmentation accuracy when objects exhibit unexpected behaviors during manipulation. 

Nadon et al. \cite{nadon2020} use scene identification during the initialization phase, which plays a crucial role in the framework, as it provides a comprehensive understanding of the workspace. By mapping out the relative positions of the robot, objects, and potential obstacles, this phase enables better task planning and object handling strategies. The ability to distinguish between rigid and non-rigid items and to adaptively plan grasps or trajectories based on scene complexity is a notable advantage, especially for mixed-object environments. Thus, this structured approach to segmentation and scene identification improves the flexibility and adaptability of robotic manipulation systems.

In other works, Henrich et al. present a promising approach to segmentation within deformable object reconstruction, though they acknowledge some limitations \cite{henrich2023}. The integrated segmentation and reconstruction model avoids the complexities of deformable registration, segmenting regions directly from point cloud data by using multi-class occupancy functions. This is particularly beneficial in scenarios like robot-assisted surgery, where accurate segmentation of an organ’s regions is crucial for precise interaction. The system leverages PointNet++ and an efficient sampling algorithm, "SortSample," to enhance the segmentation quality around object boundaries, achieving notable segmentation accuracy in both simulated and real-world data.

However, the method faces challenges, especially in real-world applications. When trained on synthetic data, the segmentation accuracy drops when applied to actual physical objects, with the model struggling to generalize fully from simulation to reality. Another limitation is the requirement for watertight meshes for the occupancy function testing, which excludes some available datasets and limits its broader applicability. Additionally, the technique’s reliance on synthetic data raises concerns about real-world robustness, particularly in complex, dynamic environments. The approach, while reducing the need for registration, may still face difficulties adapting to real-time deformations without further refinement in segmentation and tracking capabilities. Future work aims to address these issues by enhancing generalization and data diversity.

\subsubsection{Detection} 
Detection involves estimating the positions of particles that represent DOs within a single frame using preprocessed sensory input. Estimating the state of DOs from a single-frame point cloud, especially with occlusion, presents a complex challenge. Techniques like the one developed by 

Chen et al. represent progress in this area \cite{chen2024}. Their method DiPac for deformable object detection and state estimation uses particle-based modeling from RGBD images. DiPac detects deformable objects by converting depth images into point clouds, leveraging RGB data segmentation to isolate and detect object-related points, effectively filtering out the background. To improve detection accuracy, it merges multiple views into a unified 3D point cloud, constructing a complete representation and filling in occluded areas with uniformly sampled particles to model the object's unseen interior. This particle-based detection method is flexible, allowing DiPac to represent and track various deformable objects such as beans, cloth, and liquids within a single framework.

A key benefit of DiPac’s detection approach is its adaptability, particle-based detection and modeling allow for accurate, dynamic tracking, supporting gradient-based optimization in planning and control. DiPac also narrows the sim-to-real gap by fine-tuning model parameters via differentiable particle dynamics. However, this detection method relies on limited occlusion and straightforward segmentation, which may pose challenges in cluttered or complex environments. When heavy occlusions or background noise are present, accurate detection and tracking are limited. Future improvements in particle extraction under occlusion would enhance DiPac’s robustness, enabling its application in real-world unstructured environments with complex visual setups.

\subsubsection{Tracking} 
Tracking involves estimating the deformation of a DO over a sequence of frames, aiming to maintain a state consistent with the DO’s geometry in each frame. This becomes particularly challenging when parts of the object are occluded. Advances in computer vision have recently led to effective methods for reconstructing DOs in real-time, which dynamically adapt the tracking model to counteract occlusion issues \cite{newcombe2015}, \cite{dou2017}. However, these methods often lack the model consistency necessary for visual-servoing algorithms in deformable object manipulation (DOM).

DO tracking is also a focus in surgical robotics, where studies have shown significant progress in tracking soft tissues during medical procedures \cite{haouchine2013}, \cite{collins2016}. However, these approaches are usually specific to the surgical domain and may not work for other objects, like ropes or fabrics. In robotics, physics simulation is often employed to track partially occluded DOs, with several studies adopting this method \cite{schulman2013}, \cite{petit2017Pizza}. Recent approaches \cite{tang2017}, \cite{tang2018} have further used Gaussian Mixture Modeling (GMM) and Coherent Point Drift (CPD \cite{myronenko2010}) to generate inputs for physics simulations. Yet, these methods depend heavily on accurate physical models of DOs and environmental geometry, which are challenging to obtain in unstructured environments.

An important approach to tackling occlusion is Interactive perception. An example of this is introduced by Weng et al \cite{weng2024}. This paper introduces an advanced methodology for tracking deformable objects within an interactive perception framework, utilizing an active camera and a specialized Dynamic Active Vision Space (DAVS). The system is designed to tackle the complexities associated with deformable object manipulation, such as occlusions and high degrees of freedom, by integrating a dual-arm setup with a coordinated active camera. Core to this approach is the identification of key structural features, termed the Structure of Interest (SOI), which dynamically inform the camera's viewpoint selection as the object's configuration evolves.

A notable contribution is the construction of DAVS, a manifold-with-boundary representation of feasible camera actions that leverages the object’s structural features to constrain the exploration space. By formulating the active camera’s control space within this constrained subspace, DAVS significantly mitigates the space complexity of action selection; a critical limitation in high-dimensional control tasks involving deformable objects. The integration of DAVS with reinforcement learning further enables efficient perception and manipulation, as the active camera dynamically repositions based on SOI adjustments, thus achieving robust, real-time tracking.

This active perception strategy not only enhances the system’s adaptability to variable object shapes and unseen dynamics but also optimizes computational resources, presenting a scalable solution for interactive perception in deformable object manipulation. 

\subsection{Tactile Perception}

Recent advances in tactile sensing technology have sparked considerable interest in using tactile methods for deformable object manipulation (DOM). While vision primarily captures global features like shape and color, tactile sensing provides detailed, local information, such as texture and friction, which is especially useful in situations where visibility is limited.

Tactile Devices: The BioTac tactile sensor is a notable tool capable of capturing a range of sensory information similar to that of the human fingertip \cite{narang2021}. It has a durable design, featuring a rigid core for its electronics and a flexible silicone outer layer, making it both resilient and cost-effective. However, BioTac’s output requires complex signal processing and data integration, leading to higher computational demands and potential delays.

Vision-based tactile sensors represent another type of tactile technology, translating tactile information into visual data by capturing surface deformations with a camera. These sensors, offering high spatial resolution, have been effectively used in various robotic manipulation tasks. GelSight, introduced by Yuan et al. in 2017, is a prominent example of such a sensor \cite{yuan2017}. However, GelSight does not mimic certain human tactile sensations like temperature, humidity, or pain. In another development, Sundaram et al. \cite{sundaram2019} created a tactile glove equipped with 548 sensors connected by conductive wires in a piezoresistive film. This glove uses deep convolutional neural networks to analyze pressure signals for tasks like object recognition, weight estimation, and tactile pattern exploration.

Consider some DOM Applications using Tactile Sensing. Although tactile sensing is crucial for manipulation in visually occluded conditions, it is rarely the sole method used. One application is cable or fabric following; for example, She et al. \cite{she2021} developed a technique to estimate the position and friction of a cable held by a robot using GelSight-based tactile sensing. Hellman et al. \cite{hellman2017} created a real-time method using BioTac for tactile perception and decision-making in tasks like sealing a bag by tracking contours. Zheng et al. \cite{zheng2022} proposed a BioTac-based approach to analyze tactile changes during page flipping, allowing robots to learn a logical page-turning sequence. Building on the tactile glove from \cite{sundaram2019}, Zhang et al. \cite{zhang2021} extended its use to a variety of tasks involving different objects. Their model combines predictive and contrastive learning techniques to estimate the 3D coordinates of both the hand and the object based solely on touch data.

\begin{table}[H] 
\scriptsize 
\centering
\renewcommand{\arraystretch}{1.2} 
\caption{Overview of recent literature in perception for deformable object manipulation}
\label{tab:one}
\begin{tabular}{p{2.7cm} p{3.7cm} p{3.7cm} p{4.3cm}}

\hline
\textbf{Category} & \textbf{Advantages} & \textbf{Disadvantages} & \textbf{Representative Literature} \\
\hline
Visual perception & 
\begin{itemize}[noitemsep, leftmargin=0.7em]
    \item Effective prediction of deformable object states.
    \item Accurate object representation (affordance, keypoints, point clouds).
    \item High performance for specific tasks.
    \item Improved precision via advanced sensing.
\end{itemize} & 
\begin{itemize}[noitemsep, leftmargin=0.7em]
    \item Struggles with unseen scenarios.
    \item High computational and sensory demands.
    \item Limited generalization across tasks.
    \item Expensive hardware requirements.
\end{itemize} & 
\begin{itemize}[noitemsep, leftmargin=0.7em]
    \item \textit{Learning Foresightful Dense Visual Affordance for Deformable Object Manipulation} \cite{wu2023learning}
    \item \textit{Learning Visual-Based Deformable Object Rearrangement with Local Graph Neural Networks} \cite{deng2023learning}
    \item \textit{DeformerNet: Learning Bimanual Manipulation of 3D Deformable Objects} \cite{thach2024deformernet}
\end{itemize} \\
\hline
Tactile perception & 
\begin{itemize}[noitemsep, leftmargin=0.7em]
    \item Excels where vision fails (e.g., occlusion, force-sensitive tasks).
    \item Enables precise force and position control for delicate interactions.
    \item Enhances adaptability with integration into learning frameworks.
    \item Combines multimodal perception for richer object understanding.
\end{itemize} & 
\begin{itemize}[noitemsep, leftmargin=0.7em]
    \item Highly task-specific, requiring customization for new applications.
    \item Dependence on specialized tactile sensors increases cost and complexity.
    \item Computational demands for real-time performance and multimodal processing.
    \item Limited generalization across diverse tasks or environments.
\end{itemize} & 
\begin{itemize}[noitemsep, leftmargin=0.7em]
    \item \textit{Precise Robotic Needle-Threading with Tactile Perception and Reinforcement Learning} \cite{yu2023precise}
    \item \textit{VIRDO++: Real-World, Visuo-tactile Dynamics and Perception of Deformable Objects} \cite{wi2024virdo++}
    \item \textit{Interaction Control for Tool Manipulation on Deformable Objects Using Tactile Feedback} \cite{zhang2023interaction}
\end{itemize} \\
\hline
Interactive perception & 
\begin{itemize}[noitemsep, leftmargin=0.7em]
    \item Enhances understanding of deformable objects through active engagement.
    \item Improves performance via dynamic adjustments and interactive feedback.
    \item Expands manipulation capabilities using novel methods (e.g., assistive tools, markerless systems).
    \item Offers better modeling and data richness compared to static perception.
\end{itemize} & 
\begin{itemize}[noitemsep, leftmargin=0.7em]
    \item High computational and hardware demands.
    \item Task-specific approaches limit generalization.
    \item Dependence on complex setups (e.g., multi-view systems, assistive tools).
    \item Struggles with unseen objects or scenarios without extensive retraining.
\end{itemize} & 
\begin{itemize}[noitemsep, leftmargin=0.7em]
    \item \textit{Interactive Perception for Deformable Object Manipulation} \cite{weng2024}
    \item \textit{Enhancing Deformable Object Manipulation by Using Interactive Perception and Assistive Tools} \cite{zhou2023enhancing}
    \item \textit{HMDO: Markerless Multi-view Hand Manipulation Capture with Deformable Objects} \cite{xie2023hmdo}
\end{itemize} \\
\hline
\end{tabular}
\end{table}


\subsection{Summary on Perception}
In the domain of deformable object manipulation, addressing occlusions is a critical requirement for successful perception and control, as occlusions can obscure essential details needed for precise manipulation. A multi-view camera setup is instrumental in overcoming these challenges, as a single perspective often fails to capture the full structure of a deformable object in dynamic scenarios. Leveraging multiple RGB-D sensors positioned from various angles enables comprehensive 3D modeling, thus mitigating occlusion and providing a holistic representation of the scene. Nadon et al. emphasize that such a multi-view approach to segmentation allows a system to capture a deformable object’s complex configurations as they shift during manipulation, making it essential for accurate real-time robotic control in unstructured settings \cite{nadon2020}.

However, as occlusions evolve with the object's configuration throughout manipulation tasks, an active camera system that can dynamically adjust viewpoints becomes necessary. Without this adaptability, robots may lose track of key object features, especially in tasks involving high degrees of freedom. Weng et al. propose an interactive perception approach utilizing a Dynamic Active Vision Space (DAVS), allowing the camera’s viewpoint to adjust in response to changing occlusions. The active camera is coordinated with a dual-arm setup, which collectively combats occlusions as they develop, thus ensuring robust tracking and scene understanding in real-time \cite{weng2024}.

Despite advances in visual perception, some occlusions remain intractable with vision alone, especially when visibility of certain object regions is entirely obstructed. Tactile perception steps in to fill this gap, providing detailed, localized data such as texture and friction, which are crucial for successful manipulation when visual cues are lacking. Narang’s work with the BioTac sensor, which replicates human fingertip sensitivity, illustrates the unique role tactile sensing plays in deformable object manipulation under occlusion. By complementing visual information, tactile sensors enhance overall perception capabilities, allowing robots to interact with deformable objects in occluded or visually inaccessible areas \cite{narang2021}.

The necessity for rich, detailed datasets cannot be overstated when it comes to deformable object registration. In particular, highly varied datasets are required to capture the unpredictable and varied nature of deformable objects, enabling more precise registration models that can adapt to complex, real-world environments. Nadon et al. underscore the importance of these datasets, as they provide the foundational data needed to train and refine perception algorithms capable of handling the intricacies of deformable objects \cite{nadon2020}.

Additionally, whole-scene identification before planning and control is fundamental for generating a high-level understanding of the workspace, which includes mapping out object positions, obstacles, and other environmental features. This comprehensive scene awareness is critical to guiding subsequent high-level planning and effective manipulation. Nadon et al. highlight that this initial identification stage serves as a critical data-gathering step, enabling robots to distinguish between deformable and rigid objects and plan accordingly in mixed-object environments \cite{nadon2020}.

For complex tasks where both arms are engaged in manipulation, the use of a dual-arm active camera system is particularly effective in addressing occlusions. By coordinating the actions of both arms with an active camera, the system can reorient the camera as objects are manipulated, dynamically reducing occlusion challenges. This setup, advocated by Weng et al., provides enhanced adaptability and computational efficiency in scenarios involving highly variable object shapes and configurations \cite{weng2024}.

Finally, while methods such as Gaussian Mixture Modeling (GMM) and Coherent Point Drift (CPD) have proven effective for tracking deformable objects, their application in unstructured environments remains challenging. Tang et al. and Myronenko demonstrate that these methods perform well in structured settings but require adaptation for real-world environments, where varied lighting, clutter, and irregular surfaces present additional challenges. Enriching datasets with real-world variability can help these models to adapt to the complexities of deformable object tracking, allowing these techniques to function more effectively in unstructured and dynamic contexts \cite{tang2018} \cite{myronenko2010} \cite{narang2021}.

Together, these approaches create a robust framework for managing occlusions in deformable object manipulation, combining multi-perspective visual systems, active camera adjustments, tactile perception, and advanced modeling techniques to support the high-level planning and control required for effective manipulation in real-world environments. For further notable literature in perception see Table \ref{tab:one}

\begin{itemize}
  \item Multiple camera perspectives reduce occlusion issues by capturing comprehensive 3D models \cite{nadon2020}. 
  \item Active cameras, with dynamically adjusting viewpoints, are necessary to track deformable objects as they change shape \cite{weng2024}.
  \item Tactile perception addresses occlusions that are inaccessible to vision alone \cite{narang2021}.
  \item Rich, detailed datasets are essential for accurate deformable object registration in complex environments \cite{nadon2020}.
  \item Whole-scene identification prior to planning provides necessary context for manipulation tasks \cite{nadon2020}.
  \item Dual-arm active camera systems effectively combat occlusions through coordinated control \cite{weng2024}.
  \item Gaussian Mixture Modeling (GMM) and Coherent Point Drift (CPD) need real-world dataset adaptations for effective use in unstructured environments \cite{tang2018} \cite{myronenko2010} \cite{narang2021}.
\end{itemize}

\section{Modeling}

The simulation and modeling of deformable object dynamics commonly rely on geometric representations using particles or meshes \cite{muller2008} \cite{bender2017}. According to Newton's second law, the motion of a particle or vertex is governed by the time derivative of momentum and the forces applied, as described by:

\begin{equation} \label{eq:2}
M \ddot{x} = f
\end{equation}

where $M$ represents the system's mass, and \(x_i^{t+1} = x_i^t + v_i^t \Delta t\) describes the system's state. The state can either be represented by a finite set of particles (discrete case) or defined using a displacement function (continuous case). The evolution of the system's motion is determined by integrating from an initial state \(x_0\). For each particle, the progression is described as:
 
\begin{equation} \label{eq:3}
\begin{aligned}
&x_i^{t+1} = x_i^t + v_i^t \Delta t\\
&v_i^{t+1} = v_i^t + \frac{\Delta t}{m_i} \left( f_i^{\text{int}} + f_i^{\text{ext}} \right)
\end{aligned}
\end{equation}

where \(\dot{x}_i = v_i\), and the applied forces are explicitly divided into two parts. The term \(f_i^{\text{ext}}\). represents external forces, such as gravity or input forces, which are known at the current time step. For systems under constraints, this may also include external constraint forces from boundary conditions. The term \(f_i^{\text{int}} = f_i^{\text{int}}(x^t, v^t, \theta)\) describes internal forces that depend on deformation-related parameters such as the deformed configuration \(x^t\), velocity \(v^t
\), and material properties \(\theta\). Properly modeling these internal forces is crucial for simulating deformable objects.

Once the force terms are determined, the simulation can proceed using explicit Euler integration, similar to the scheme in equation \ref{eq:3}. Other methods, such as semi-implicit integration or Runge-Kutta methods, may be used for improved accuracy. However, stable simulations often require small time steps, where the step size depends on the specific application. A more robust approach is implicit integration, such as the backward Euler method. In this case, the internal force \(f_i^{\text{int}} = f_i^{\text{int}}(x, v, \theta)\) depends on future states \(x^{t+1}\) and \(v^{t+1}\) making the integration of equation \ref{eq:3} less straightforward, as it requires solving for \(x^{t+1}\) and \(v^{t+1}\).

The advantage of implicit integration lies in its stability since the subsequent states remain closer to the dynamic constraints. Specific simulation methods often vary in their choice of principles and integration schemes, balancing between accuracy, stability, and computational cost. For instance, in computer graphics applications like character or scene animation, precise offline simulations are often acceptable, allowing for careful parameter tuning. In robotics, however, robustness and computational efficiency are critical for real-time tracking, interactive control, or learning in simulated environments. Additionally, accuracy becomes paramount in safety-critical fields such as medical simulations. With these considerations in mind, the next section provides a brief overview of methods based on different physical models and simulators.

\subsection{Mass-Spring Models}

\begin{figure}[ht]
    \centering
    \setlength{\fboxsep}{5pt} 
    \setlength{\fboxrule}{1pt} 
    \fcolorbox{lightgray}{white}{
        \begin{overpic}[width=\textwidth, trim={0 0 27 0}, clip]{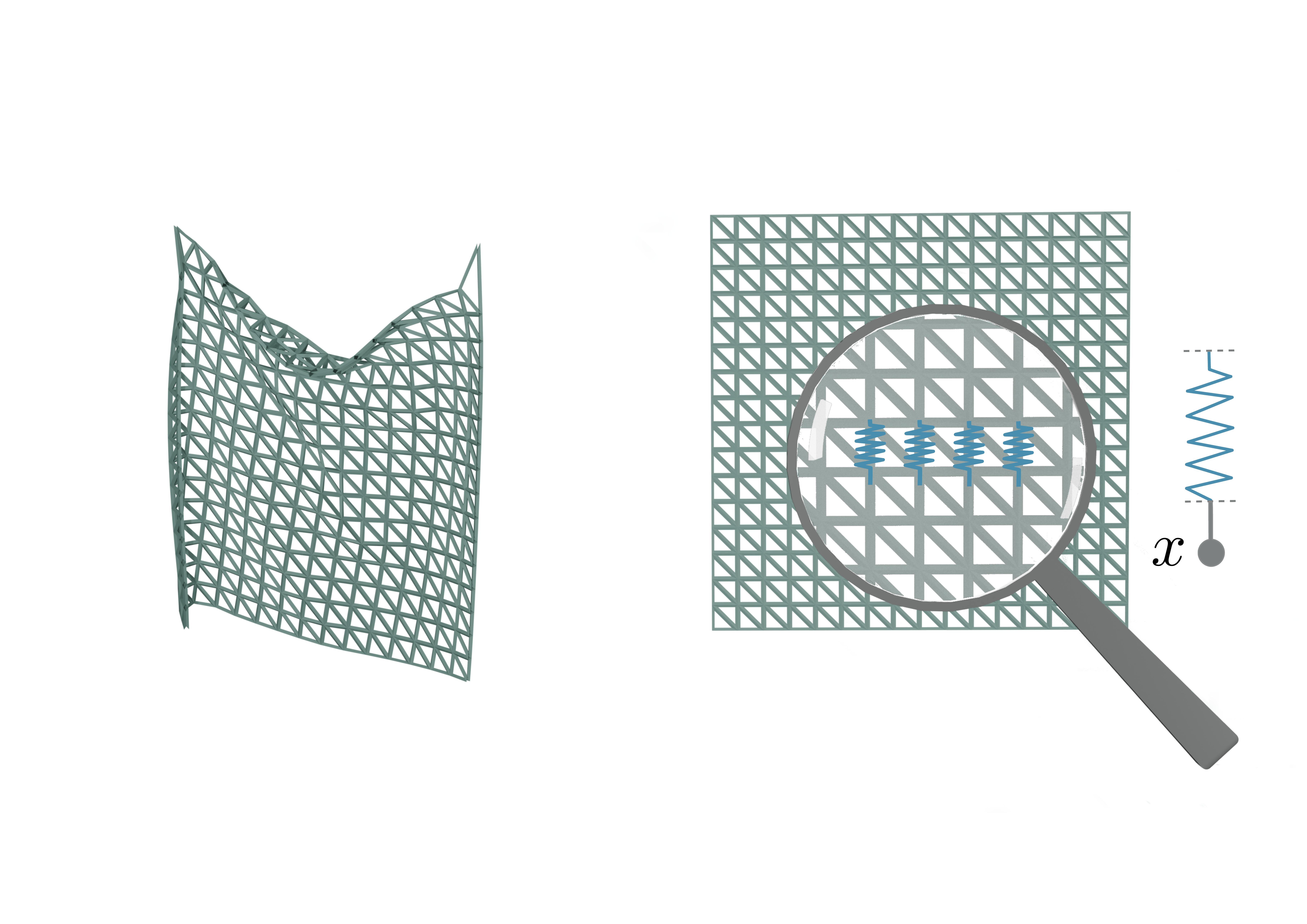}
            \put(50,5){\color{black} \large $\mathbf{F = -k x}$}
        \end{overpic}
    }
    \caption{Illustration of cloth modeling using a mass-spring system: a triangular mesh structure representing the cloth is depicted, with a magnified view showing the individual mass points and spring connections, highlighting the deformation mechanics and the use of Hooke's law (bottom centre). }
    \label{fig:mass-spring-system}
\end{figure}

Mass-spring systems (MSSs) represent deformable materials as a network, where each vertex is assigned a mass, and vertices are connected by spring edges. The internal force between two vertices, $i$ and $j$, is calculated based on a spring-damping relationship along the direction of displacement. In implicit integration, these equality constraints are linearized around the current state, enabling the use of methods such as Newton-Raphson to solve the system \cite{muller2008}. MSSs are straightforward to implement and offer fast simulation speeds. They have been used to simulate various deformable objects, like modeling a rope as a chain \cite{schulman2013case} or a cloth as a two-dimensional (2D) grid \cite{kita2011}. However, a key limitation is that MSSs are most effective for small deformations and struggle with accurately capturing complex elastic behaviors. Even increasing mesh resolution doesn’t ensure convergence to the correct behavior, as the network's topology and spring parameters also play a role \cite{muller2008}. Moreover, the spring parameters lack a direct correlation to material properties, requiring significant tuning to achieve the desired dynamic behavior. For an illustration of this concept see Figure \ref{fig:mass-spring-system}.

\subsection{Position-based-dynamics}

\textit{Position-based dynamics (PBD)} \cite{muller2008} is a meshless approach that models materials as a discrete system of particles. The simulation uses an implicit integration scheme where internal forces are derived from holonomic constraints, including temporary or unilateral constraints. For an illustration of this concept see Figure \ref{fig:mass-spring-system} however note rather than merely using hook's law PBD allows for the use of a broad range of relations and boudary conditions.

The update equations are given by:

\begin{equation} \label{eq:4}
\begin{aligned}
x_i^{t+1} & =x_i^t+\Delta t v_i^{t+1} \\
v_i^{t+1} & =v_i^t+\frac{\Delta t}{m_i}((f_i^{\mathrm{xt}}-\sum_j k_j \nabla_{x_i^{t+1}} C_j\left(x_1^{t+1}, \ldots, x_n^{t+1}\right)))
\end{aligned}
\end{equation}

Here, \(k_j\) represents the stiffness of the \(j\)th constraint \(C_j(x_1, x_2, \dots, x_n) = 0\), where \(j = 1, 2, \dots\). Unlike mass-spring systems (MSSs), PBD directly updates the position \(x_i^{t+1}\) before calculating the velocity using $v_i^{t+1} = (x_i^{t+1} - x_i^t) / \Delta t$.

As noted in \cite{bender2017}, the future state \(x_i^{t+1}\) solves an optimization problem constrained by Equation \eqref{eq:4}, where \(k_j \to \infty\). PBD solves this optimization approximately by iteratively projecting the candidate position \(\hat{x}_i^{t+1}\) onto constraint manifolds, often using a Gauss-Seidel method. The elastic properties depend on the constraint parameters \(\theta\), the strictness of enforcement (via \(k_j\)), and the number of projection iterations. 
 
Numerous constraints have been proposed to simulate various effects, such as volume conservation, frictional contact, or strain under continuum mechanics \cite{bender2017}. 

PBD offers fast, stable simulations with full control, enabling the easy incorporation of constraints and boundary conditions. The method is also versatile, capable of modeling plastic deformation, fluids, and rigid-body dynamics \cite{macklin2014}. In robotics, PBD can be used to simulate manipulators, deformable objects, and their interactions within a unified framework. However, methods based on spring forces struggle with stiff systems like articulated rigid bodies. 

A key drawback of PBD is that it does not simulate forces accurately, and system stiffness is tied to the time-step size. A solution using constraint compliance has been suggested in more recent extensions \cite{macklin2016}. Still, PBD typically relies on impulse updates, producing visually plausible but not always physically accurate simulations. Additionally, it can be challenging to interpret some parameters physically. While constraints between particles are intuitive, linking simulation results to physical material properties, like modulus, is less straightforward. This often requires tuning to achieve desired effects, and parameter identification with PBD can only yield qualitative material characterizations \cite{guler2015}, similar to MSSs.

\subsection{Continuum Mechanics}

Continuum mechanics provides a more accurate physical model for describing material deformation in a continuous domain. The state at time \( x_t \) represents a displacement function over material coordinates. To describe material deformation, displacement is typically measured relative to an initial configuration \( x_0 \), known as the rest shape. The gradient of this displacement field reveals the distortion of the original element geometries. As a result, the conservation of momentum Equation \eqref{eq:5} applies to each element of the domain:

\begin{equation} \label{eq:5}
\rho \ddot{x} = f_{\text{int}} + f_{\text{ext}} = \nabla \cdot \sigma + f_{\text{ext}}
\end{equation}

where \( \rho \) represents density, and the force terms correspond to body forces applied to a unit volume. The symbol \( \sigma \) denotes the stress tensor, which is a symmetric matrix. Taking the dot product with the gradient operator, the stress term captures internal elastic effects. The relationship between stress and material deformation is defined by a constitutive model. For Hookean materials, this relationship is linear as seen in Equation \eqref{eq:6}.

\begin{equation} \label{eq:6}
\sigma = E \epsilon
\end{equation}

where \( E \) is a coefficient matrix (e.g., 6 × 6 for the 3D case). For isotropic materials, this matrix is fully determined by two scalars: (i) Young’s modulus, which measures material stiffness, and (ii) Poisson’s ratio, which is the ratio of lateral to longitudinal strain. The strain tensor \( \epsilon \) is also a symmetric matrix, dependent on the deformation gradient of the displacement field:

\begin{equation} \label{eq:7}
\epsilon = \nabla p + (\nabla p)^T + (\nabla p)^T \nabla p
\end{equation}

where \( p = x - x_0 \), known as the Green-Lagrange strain. A simplified version of this, called Cauchy strain, ignores the last quadratic term to yield a linear relationship with the displacement, assuming geometric linearity. The Hookean material assumption is typically restricted to modeling small deformations around the rest shape. More advanced constitutive models include energy terms that penalize deformation, leading to nonlinear hyperelastic models like the Neo-Hookean model \cite{boonvisut2013}. Simulating such systems involves solving the partial differential equation \eqref{eq:5}. The original form of the equation can be solved effectively only for simple, small-scale problems, such as a 1D string. For more complex cases, the finite element method (FEM) is the standard computational approach for solving continuum mechanics problems.

FEM works by discretizing the material domain into a mesh of finite geometry elements, such as tetrahedra. Global consistency is enforced through constraints on element boundaries, and the physical fields are approximated as a linear combination of basis functions. Within each element, a weak form of the governing equation can be transformed into an ordinary differential equation, which is easier to solve via numerical integration. Similar to mass-spring systems (MSSs), implicit integration requires solving the internal force term through linearization at each time step, which can be complex in FEM when using a general constitutive model.

Linear FEM assumes both material and geometric linearity, enabling offline precomputation for efficient simulation. However, the Cauchy linear strain provides a poor approximation for the rotational component of the displacement field. Corotational linear FEM addresses this by extracting the rigid rotation matrix and incorporating it into the stress-strain relationship, compensating for the rotational component.

\begin{figure}[htbp]
    \centering
    \setlength{\fboxsep}{4pt}  
    \setlength{\fboxrule}{0.8pt}  
    \fcolorbox{lightgray}{white}{%
        \includegraphics[width=0.95\linewidth, trim=0 0 0 0, clip]{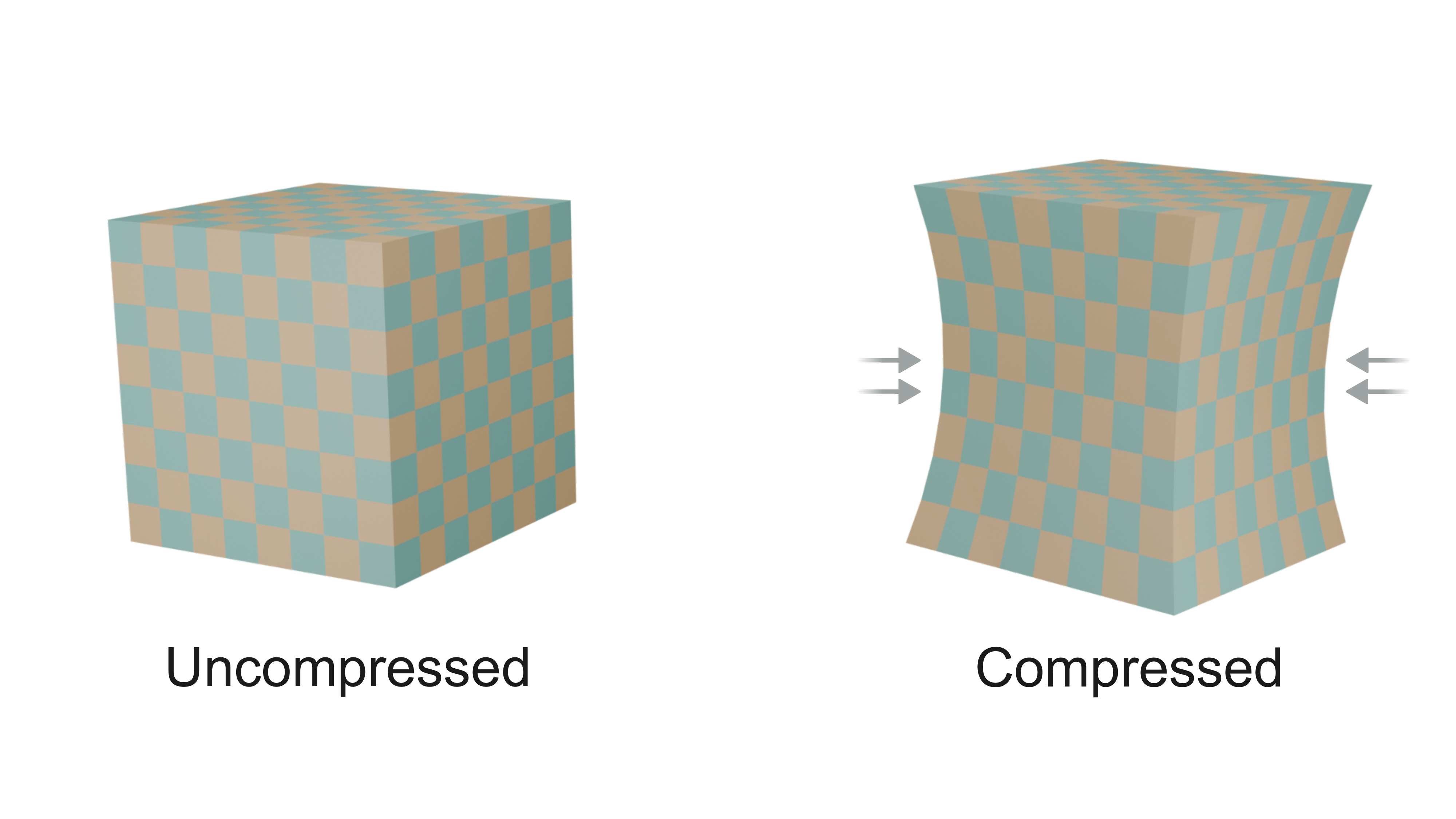}
    }
    \caption{The undeformed cube (left) represents the reference configuration, while the deformed cube (right) illustrates the effect of compressive forces acting along the horizontal sides (grey arrows). The deformation is described by the deformation gradient (\(\mathbf{F}\)), where \(\mathbf{F} = \frac{\partial \mathbf{x}}{\partial \mathbf{X}}\), mapping the reference configuration (\(\mathbf{X}\)) to the deformed configuration (\(\mathbf{x}\)). The strain in the material is captured by the strain tensor \(\boldsymbol{\varepsilon} = \frac{1}{2} \left( \nabla \mathbf{u} + (\nabla \mathbf{u})^\top \right)\), with \(\varepsilon_{xx} < 0\) indicating compression along the horizontal axis. The stress tensor (\(\boldsymbol{\sigma}\)) satisfies the equilibrium condition \(\nabla \cdot \boldsymbol{\sigma} = 0\) under static loading, and the material behavior follows Hooke’s law: \(\boldsymbol{\sigma} = \mathbf{C} : \boldsymbol{\varepsilon}\). The top and bottom faces remain undeformed due to fixed boundary conditions (\(u_y = 0\)).}
    \label{fig:continuum-mechanics}
\end{figure}

Modeling deformable objects using continuum mechanics leads to more accurate simulations, which is essential in applications where precision is critical. The parameters of the model have clear physical interpretations. Frameworks like FEM are widely used for their versatility in simulating various deformation effects. However, these computations and implementations are more complex compared to particle-based methods. Linear FEM is commonly used in robotics when accuracy under large deformation is not a priority. Nonlinear computational models generally struggle to achieve real-time performance, except in cases where explicit integration or hardware acceleration \cite{zollhofer2014} is employed. For an illustration of this concept see Figure \ref{fig:continuum-mechanics}.

\subsection{Physics-based Simulation}

Several simulators have been developed based on the theoretical foundation mentioned earlier, offering development environments for communities in graphics, computer vision, robotics, and control. Here, we briefly discuss some of the most commonly used simulators, emphasizing their modeling techniques and applications in robotics. The Simulation Open Framework Architecture (SOFA) is an open-source framework designed for tissue modeling and surgical interaction \cite{faure2012}. It has a modular design that accommodates customized solvers for various mechanical objects, constraints, and collision geometries. SOFA includes built-in mass-spring models and a variety of finite element method (FEM) models, such as linear, corotational linear, and Neo-Hookean models, which enhance precision in medical applications. SOFA has been utilized in areas like motion planning \cite{yoshida2015}, tracking deformable objects \cite{matas2018},\cite{mcconachie2018} and vision-based tip force estimation \cite{petit2017}, \cite{haouchine2018}. 

A key strength of SOFA lies in its \textbf{open-source and modular architecture}, distributed under the LGPL license, which empowers users to extend or reconfigure nearly every element of the simulation engine—from solvers to collision pipelines, mappings, and force models—without being locked into closed internals \cite{faure2012sofa}. This design enables \textit{complete control over the physics stack}, such that one can substitute or fine-tune integration schemes, linear solvers, constraint resolution strategies, and coupling mappings to precisely tailor the simulation to one’s scenario \cite{faure2012sofa,sofa2023performance}. In particular, researchers in soft robotics have leveraged this flexibility via plugins like \textit{SoftRobots}, enabling real-time direct/inverse FEM modeling, custom actuation, and coupling between rigid and deformable elements \cite{duriez2018softrobots}. Because SOFA supports multiple constitutive models (e.g., linear, corotational, Neo-Hookean), selective solver pipelines, and model order reduction or GPU offloading, it consistently delivers \textbf{state-of-the-art deformable object performance} in academic use cases \cite{faure2012sofa,sofa2023performance}. 

Importantly, despite its high fidelity, SOFA is engineered to maintain \textbf{efficient computation even on moderate hardware}. Its architecture allows parallel execution of collision detection and free motion, asynchronous solver execution, block-sparse assembly, constant sparsity pattern reuse, and optional GPU acceleration via CUDA plugins \cite{sofa2023performance}. Users can trade off solver tolerances, iteration limits, or exploit reduced-order modeling to reduce cost while preserving essential deformation fidelity. In practice, many published works report complex soft robotic simulations (with full coupling to rigid elements) running in real time on desktop or workstation-class machines without requiring ultra–high-end GPU clusters \cite{nunes2024ppo}. This balance of high-fidelity deformable simulation, full user control, and computational efficiency makes SOFA particularly attractive for research in soft robotics, surgical interaction, and deformable manipulation, where one often must integrate novel physical models, sensors, and control without being constrained by ``black-box'' simulation limits.

PhysX, developed by NVIDIA, is another open-source simulator widely used in game engines. It incorporates cloth modeling based on Position-Based Dynamics (PBD) \cite{nvidia2019}, allowing the simulation of both rigid and flexible objects without requiring multiple simulators, as seen in \cite{bai2016}. With extensions for handling frictional contact, PhysX has been employed in human dressing tasks, including haptics and force prediction \cite{yu2017}, \cite{erickson2018}, and as a forward dynamics model for planning dressing motion \cite{kapusta2019}.

Another simulator based on PhysX; Isaac Gym \cite{makoviychuk2021} offers groundbreaking modeling techniques and simulation capabilities that enhance robotic manipulation of deformable objects. By leveraging NVIDIA’s high-performance PhysX engine, Isaac Gym allows continuous control tasks to run directly on the GPU, achieving exceptional simulation speeds without the need for CPU-based physics calculations. This architecture enables highly parallelized simulations where thousands of environments run simultaneously, thus speeding up training by orders of magnitude compared to traditional methods.

For deformable object manipulation, Isaac Gym provides tools that can realistically model soft-body physics and complex contact dynamics. Its end-to-end GPU simulation pipeline directly integrates with PyTorch, allowing for real-time feedback between the physics engine and the neural network policy training. This integration enables learning complex control strategies for deformable objects, which typically involve continuous changes in shape and require intricate contact handling.

The platform supports diverse control environments, where robots like Shadow Hand and Allegro are trained to manipulate objects with varying textures and consistencies, advancing tasks such as squeezing, stretching, and shape adaptation. This high fidelity and performance make Isaac Gym well-suited for developing robotic applications in areas requiring soft-material manipulation, such as medical robotics, food processing, and advanced manufacturing, ultimately bridging simulation and real-world applicability in deformable object handling.

Huang et al. introduces DefGraspSim \cite{huang2021}, a physics-based simulation tool for robotic grasping of 3D deformable objects, utilizing the Isaac Gym environment with a GPU-accelerated finite element method (FEM). This model accurately represents how deformable objects respond to robotic manipulation, accounting for complex deformations, stresses, and instabilities. DefGraspSim models each object as a volumetric mesh, applying co-rotational linear FEM to capture nuanced deformation behavior under varied grasp forces. Coupled with an isotropic Coulomb friction contact model, it enables realistic simulation of grasp scenarios.

Although DefGraspSim itself is not gradient-based, it produces extensive data—over a million grasp measurements—providing a foundation for training gradient-based learning models. This dataset can support machine learning techniques, such as deep learning or reinforcement learning, which rely on gradient descent. DefGraspSim’s output aids in developing predictive grasp models and optimizing grasp strategies, making it valuable for improving robotic performance in real-world deformable object manipulation.

Another well-known simulator in this area; MuJoCo, previously proprietary engine that has subsequently been open-sourced, has been used for modeling rigid-linked characters \cite{mordatch2012} and as a backend for reinforcement learning (RL) benchmarks \cite{brockman2016}. Featuring a convex soft contact model, MuJoCo enables optimization methods with faster convergence rates \cite{todorov2014}. It takes a constraint-based approach to solving interaction forces, bearing some resemblance to PBD, and can model deformable objects. Initially designed for modeling articulated robots, MuJoCo provides useful interfaces for robot modeling and learning algorithms. Its deformable dynamics modeling has been applied in tasks like cloth folding \cite{petrik2019} and rope manipulation planning \cite{yan2020}. Lastly, Bullet is an open-source library for collision detection and multibody simulation \cite{coumans2016}. It also uses PBD to simulate deformable bodies and their interactions with other entities. Recent updates have added models for corotational and Neo-Hookean materials. Bullet offers user-friendly interfaces for robot modeling and machine learning \cite{coumans2016}. It has been used in research for various deformable dynamics tasks, including object tracking (\cite{schulman2013}, \cite{elbrechter2012}), sim-to-real transfer \cite{matas2018}, learning manipulation \cite{mcconachie2018}, and assistive dressing simulations \cite{erickson2020}.


\begin{table}[H]
\scriptsize 
\centering
\renewcommand{\arraystretch}{1.8} 
\caption{Overview of recent literature in modelling for deformable object manipulation}
\label{tab:two}
\begin{tabular}{p{3cm} p{5cm} p{5cm} p{2.5cm}}

\hline
\textbf{Category} & \textbf{Advantages} & \textbf{Disadvantages} & \textbf{Literature} \\
\hline
\textbf{Mass-Spring-Systems} & 
\begin{itemize}[noitemsep, leftmargin=0.7em]
    \item Flexible for modeling deformable objects.
    \item Accounts for material variations like stiffness and knots.
    \item Supports real-world tasks (e.g., string handling, dual-arm coordination).
    \item Learning integration improves adaptability.
    \item Ensures grasp stability in deformable object manipulation.
    \item Easy to implement in robotic systems.
\end{itemize} & 
\begin{itemize}[noitemsep, leftmargin=0.7em]
    \item High computational complexity for real-time applications.
    \item Simplified models may miss nonlinear deformations.
    \item Requires pre-computation and parameter tuning.
    \item Limited representation for complex materials (e.g., anisotropy).
    \item Needs extensive training data.
\end{itemize} & 
\begin{itemize}[noitemsep, leftmargin=0.7em]
    \item Tabata et al. (\cite{Tabata2023})
    \item Shi et al. (\cite{Shi2024})
    \item Zaidi et al. (\cite{Zaidi2020})
\end{itemize} \\
\hline
\textbf{Position-Based-Dynamics} & 
\begin{itemize}[noitemsep, leftmargin=0.7em]
    \item Efficient, stable simulation for deformable objects.
    \item Realistic modeling for ropes and DLOs.
    \item Suitable for real-world tasks (e.g., suturing, obstacle avoidance).
    \item Supports optimization via differentiable frameworks.
    \item Validated in practical scenarios.
\end{itemize} & 
\begin{itemize}[noitemsep, leftmargin=0.7em]
    \item Limited to linear or rope-like objects, reducing generalization.
    \item Computational challenges for real-time tasks.
    \item Relies on accurate parameter tuning.
    \item Sensitive to parameter estimation errors.
\end{itemize} & 
\begin{itemize}[noitemsep, leftmargin=0.7em]
    \item Yu et al.\cite{Yu2023}
    \item Liu et al.\cite{Liu2023}
\end{itemize} \\
\hline
\textbf{Continuum Mechanics} & 
\begin{itemize}[noitemsep, leftmargin=0.7em]
    \item High accuracy for modeling deformations.
    \item Provides physics-based realism.
    \item Enables blind manipulation using force sensing.
    \item Effective for grasping tasks with deformable objects.
    \item Validated through simulation and real-world applications.
\end{itemize} & 
\begin{itemize}[noitemsep, leftmargin=0.7em]
    \item Computationally expensive, limiting real-time use.
    \item Requires precise material properties and meshing.
    \item Difficult to scale for complex objects.
    \item Limited generalization to other deformable manipulation scenarios beyond the specific tasks studied.
\end{itemize} & 
\begin{itemize}[noitemsep, leftmargin=0.7em]
    \item Sanchez et al.\cite{Sanchez2020}
    \item Huang et al.\cite{huang2021}
\end{itemize} \\
\hline
\end{tabular}
\end{table}


\subsection{Summary on Modeling}
In the realm of deformable object manipulation (DOM), high-quality, real-time physical simulation is indispensable. Given the vast complexity and high degrees of freedom inherent to deformable objects, achieving precise control and manipulation requires simulators that can accurately capture intricate dynamics. Unlike rigid objects, deformable materials respond to even minor forces with nuanced, non-linear behaviors that traditional models often struggle to replicate. Advanced physics-based simulators, such as Isaac Gym and SOFA, have thus become essential. By modeling soft-body physics, complex contact interactions, and shape changes in real-time, these simulators allow developers to experiment and refine control strategies in a realistic, dynamic environment.

Further enhancing their value, these simulations integrate differentiable frameworks, providing critical, fine-grained gradients for optimization-based learning. For instance, differentiable simulators like DiffSRL empower learning algorithms to adjust with high precision, as they can access direct feedback on physical constraints. Such access is especially beneficial for reinforcement learning, enabling faster and more reliable policy training. Without these rich, physically informed insights, robotic systems often face limitations in responsiveness and adaptability, especially when transitioning from controlled simulations to unpredictable real-world environments. In summary, high-fidelity, real-time physical simulations are not just helpful; they are vital for the development of effective, adaptable DOM frameworks. For those looking for further recent notable works in modeling, please see Table \ref{tab:two}.

\begin{itemize}
  \item Real-time simulation vital for DOM
  \item Physical Simulation vital for DOM
  \item Differentiability of Simulation vital for learning algorithms
\end{itemize}

\section{Manipulation}

The purpose of manipulating deformable objects (DOs) is to find the ideal force or movement at key points to achieve a specific task. To start, we review approaches for planning and controlling deformable objects analytically, then focus on learning-based methods such as Reinforcement Learning (RL) and Imitation Learning (IL).

\subsection{Planning for DOM} Planning for DOM involves determining the optimal sequence of configurations for a robot or object to achieve the desired result. For instance, bending a rope into a particular shape can be framed as an optimization problem:

\begin{equation} \label{eq:8}
\begin{gathered}
x_{0: T}^*, u_{0: T-1}^*=\underset{x_{0: T}, u_{0: T-1}}{\arg \min } \mathcal{J}_{u_{0: T-1}}\left(x_{0: T}\right) \\
x_{t+1}=\text { SystemDynamics }\left(x_t, u_t\right)
\end{gathered}
\end{equation}

Where \( x_{0:T} \) represents the sequence of system states (e.g., manipulated degrees of freedom), and \( \mathcal{J}(\cdot) \) denotes the cost associated with executing the planned trajectory \( x_{0:T}, u_{0:T-1} \) over the time horizon \( T \).

\subsubsection{Shooting in the action  space} 
\begin{figure} 
    \centering
    \includegraphics[width=\textwidth]{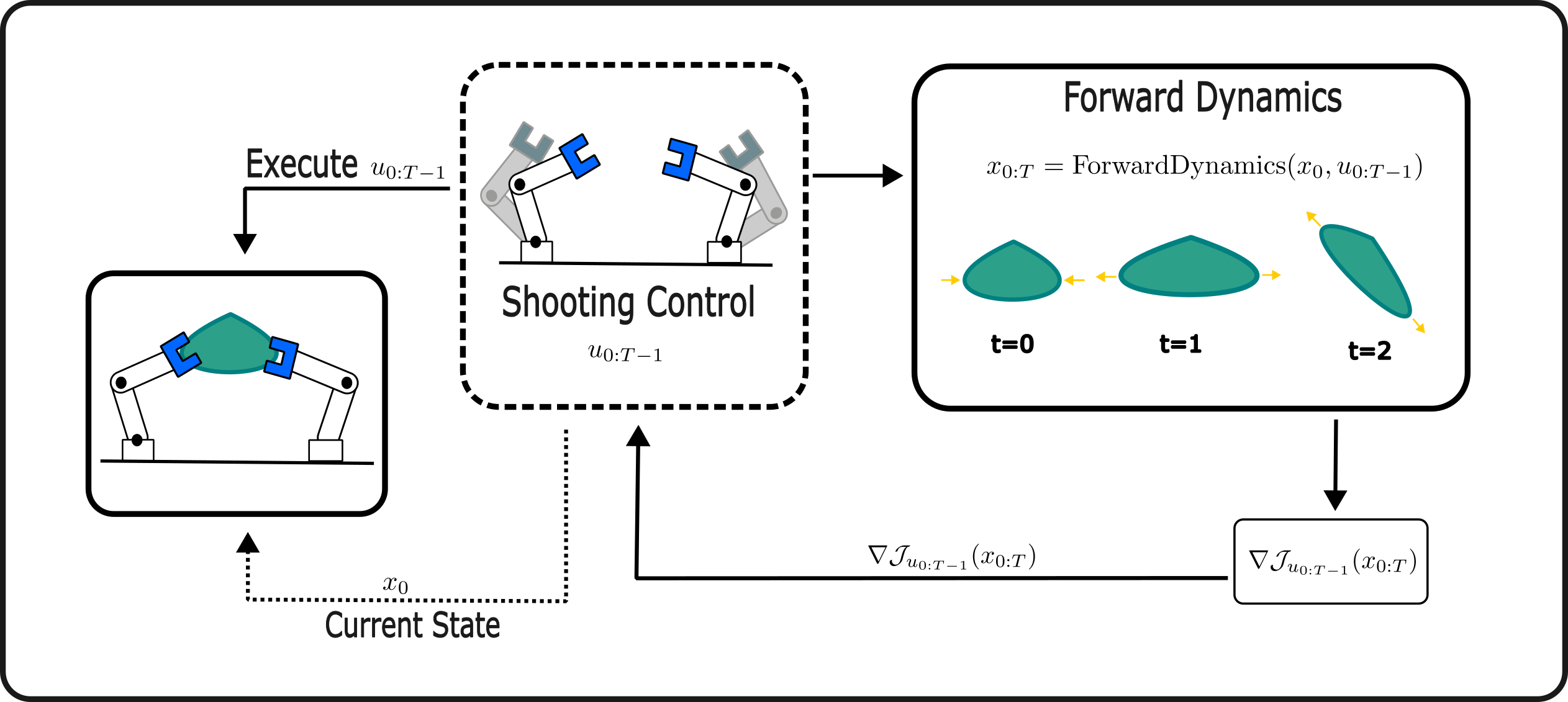}
    \caption{Shooting control diagram demonstrating forward dynamics and optimization. The constrained variables are in the action and control space (dashed  rectangle). Diagram is based on \cite{shiach2024}}
    \label{fig:shooting_control}
\end{figure}

One solution for the above optimization is to try a sample sequence of actions, then adjust based on calculated costs of projected paths. Since action space has a lower dimensionality than object state space, this method can simplify the search process. Precise dynamics modeling and efficient backward gradient evaluation are key to this approach. Huang et al. \cite{huang2023} demonstrated shape control for deformable linear objects (DLOs) using dual-arm robots in the action space, while Shi et al. \cite{shi2022} developed a method to shape Play-Doh into letters with human-like skill. Similar techniques are explored in \cite{zaidi2017}, \cite{li2015}, \cite{lin2015}. See Figure \ref{fig:shooting_control} for an illustration of this concept.

\subsubsection{Searching state trajectories}
\begin{figure} 
    \centering
    \includegraphics[width=\textwidth]{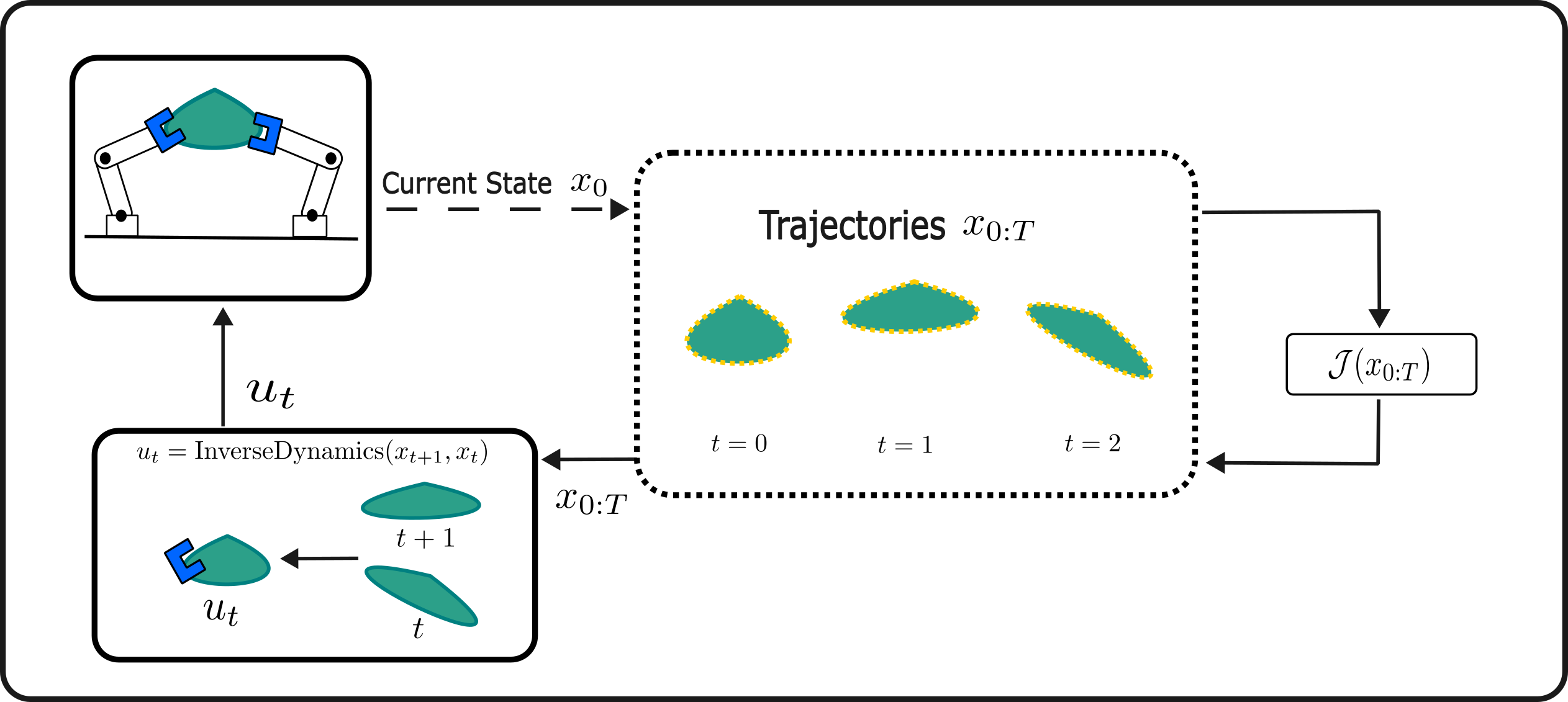}
    \caption{State trajectory exploration: The constrained variables lie within the state space (dashed rectangle), with actions determined based on state trajectories and inverse dynamics. This is a common framework in Model-Predictive-Control, a thorough explanation can be found here \cite{tedrake_underactuated}}.
    \label{fig:trajectory-search}
\end{figure}

Another approach is to search within the object’s state space to find a feasible path \( x_t \), then derive actions for each transition. This approach involves efficiently sampling configurations in a large state space. Sintov et al. \cite{sintov2020} found that the configuration space of elastic rods forms a smooth, finite-dimensional manifold and developed a sampling-based algorithm to generate feasible paths. Lui and Saxena \cite{lui2013} improved DLO shape planning by using an energy model to create a rough path that ensures task success, followed by a local controller to adjust the path with real-time feedback for accuracy. Related studies \cite{mcconachie2020}, \cite{bretl2014}, \cite{shah2016} also work on efficient state representation and transition modeling to plan state trajectories. See Figure \ref{fig:trajectory-search} for an illustration of this concept.

\subsubsection{Control Strategies for DOM}
Control aims to design the necessary inputs for robots to perform the desired motions. Closed-loop control, which uses sensory feedback, can handle various uncertainties in DOM. However, because of DOs' complex dynamics, synthesizing a global control strategy with guaranteed success is challenging. Instead, local controllers are typically used to apply force or velocity at specific points. The control action can be described as:

\begin{equation} \label{eq:9}
u_t=u\left(\phi\left(x_t\right)-\phi\left(x_g\right), \theta\right)
\end{equation}

where \( x_t \) is the current state, \( x_g \) is the goal state and \( \phi(\cdot) \) represents the feature space. Control then works to minimize the difference between the current state \( x_i \) and the target state \( x_g \), potentially within the feature space \( \phi(\cdot) \).

\begin{figure} 
    \centering
    \includegraphics[width=\textwidth]{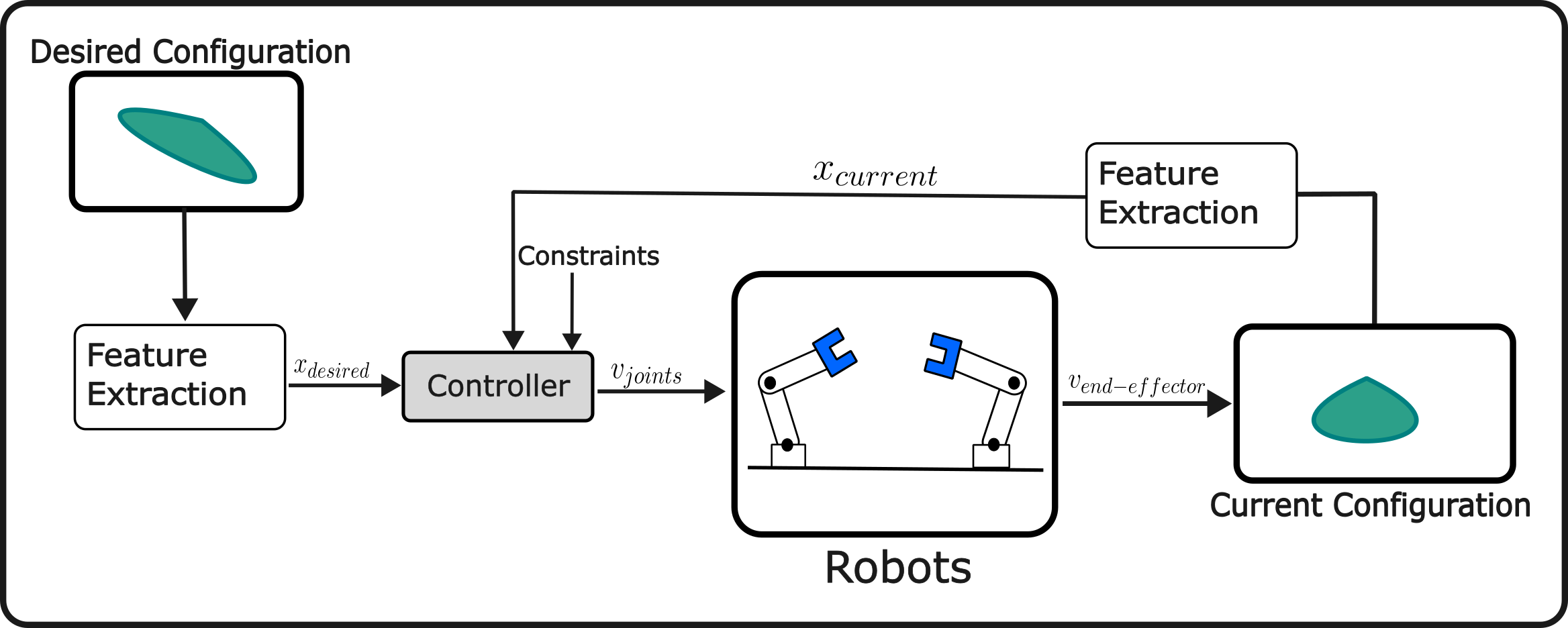}
    \caption{A basic closed-loop control framework based on visual features. A thorough description can be found here \cite{spong2005}}
    \label{fig:simple-control}
\end{figure}

Due to real-time requirements, the function \( u(\cdot) \) is typically calculated using straightforward approaches like a linear function between operating points and feature points or through single-step horizon numerical optimization methods. Some approaches rely on visual-servoing frameworks to regulate velocity, such as arranging cloth  or handling deformable linear objects (DLOs). In contrast, tactile feedback is sometimes used to refine actions, as seen in cable manipulation tasks. See Figure \ref{fig:simple-control} for an illustration of this concept.

\subsection{Learning-Based Approaches}
Learning-based approaches, particularly Reinforcement Learning (RL) and Imitation Learning (IL), offer effective ways to manage DOM tasks without needing a complex dynamical model, which can be challenging to obtain for many DOs.

\subsubsection{Reinforcement Learning for DOM} 
Reinforcement Learning (RL) operates on a trial-and-error basis, aiming to optimize control policies by maximizing accumulated rewards. The optimal policy \( \pi \) is found by maximizing the expected future cumulative reward, represented as:

\begin{equation} \label{eq:10}
\pi^*=\underset{\pi \in \Pi}{\arg \max } \mathbb{E}_{\tau \in d^\pi}\left(\sum_{t=1}^T \gamma^{t-1} r_t\right)
\end{equation}

\begin{figure} 
    \centering
    \includegraphics[width=\textwidth]{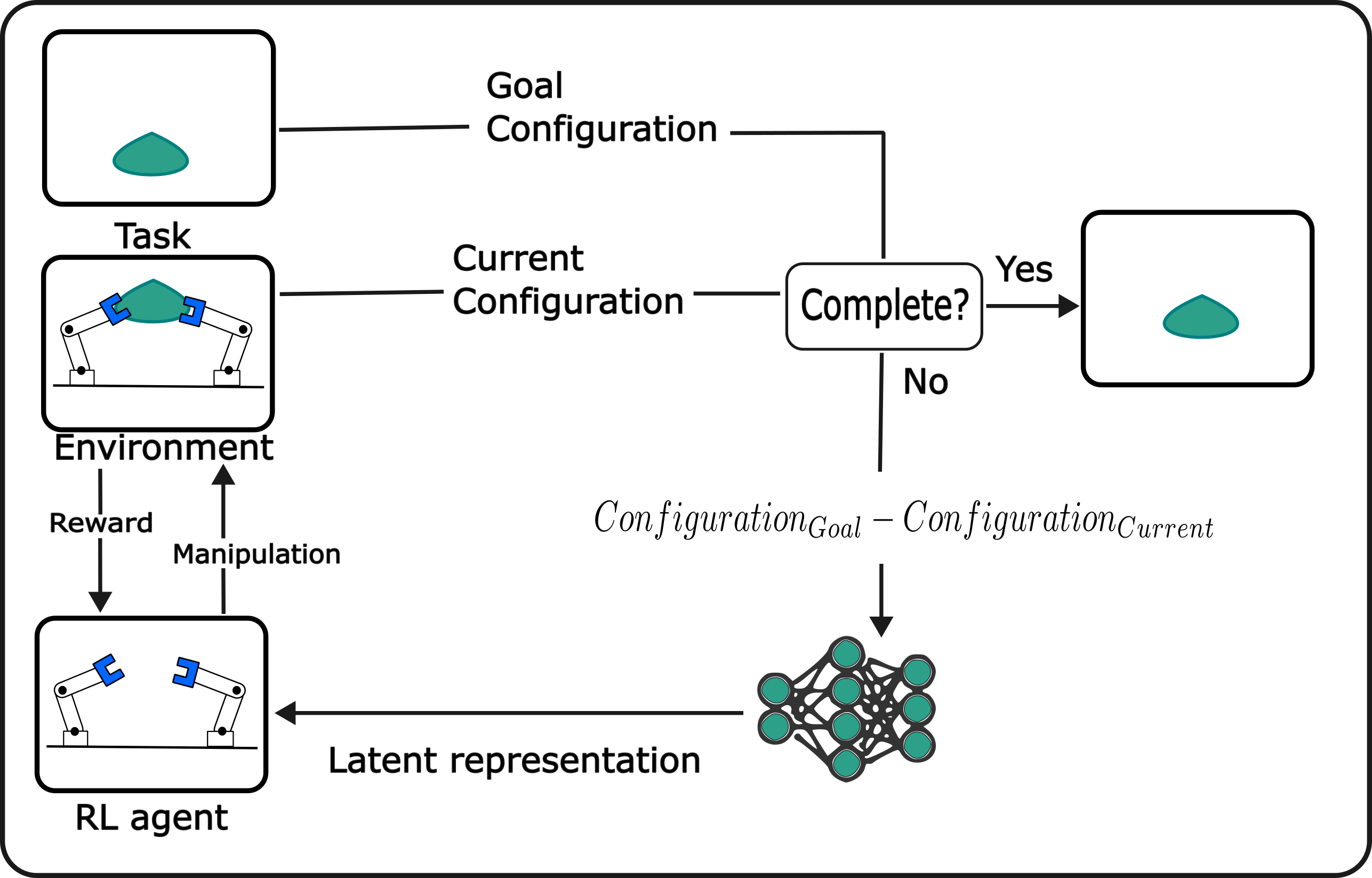}
    \caption{RL pipeline: First, the configurations of the DO are encoded into a latent space. Next, the RL agent generates manipulation actions by mapping the current and goal latent representations, enabling continuous interaction with the environment until the task is successfully completed. A thorough introduction to these concepts can be found in \cite{sutton2018}}
    \label{fig:rl-pipeline}
\end{figure}

Here, \( \tau \) represents a trajectory from the policy's trajectory distribution \( d^{\pi} \) of trajectories generated by the Markov-Decision-Process (MDP) or policy. $r_t$ is the reward at step \( t \) and \( \gamma \) is a discount factor (0 to 1) that moderates the impact of future rewards. Model-free RL, which doesn’t require an explicit model of DO dynamics, faces challenges like sample inefficiency, especially with infinite-dimensional DOs. Additionally, extensive trial-and-error in real environments can be risky. Simulators help address these challenges by allowing initial training in simulated settings. The main obstacle then is translating these learned strategies to real robots. Embedding structures or reducing feature dimensions can enhance RL efficiency. See Figure \ref{fig:rl-pipeline} for an illustration of this concept.

Chen et al. introduces DaXBench \cite{chen2022daxbench} in which reinforcement learning (RL) is enhanced with differentiable physics, providing a notable improvement in control and efficiency over traditional methods. The Short-Horizon Actor-Critic (SHAC) and Analytic Policy Gradients (APG) methods excel in low-level control tasks. For example, in the Whip-Rope task, APG and SHAC achieved near-optimal performance, with APG reaching a reward of 1.00 (perfect performance), significantly outperforming Proximal Policy Optimization (PPO), which achieved a reward of only 0.34.

In high-level planning tasks, expert demonstrations further reduce exploration and sample inefficiency, guiding RL agents toward productive trajectories. This effect is critical in complex, sparse-reward tasks like cloth folding, where random exploration can be highly inefficient. For example, with expert demonstration integration, Imitation Learning via Differentiable Physics (ILD) achieved a reward of 0.85 in the Fold-T-shirt task. 

DaXBench’s differentiable physics also enables generalization across deformable object types by allowing RL agents to use analytical gradients to navigate diverse manipulation tasks. In Pour-Water, SHAC's reward reached 0.91, compared to PPO's 0.32, demonstrating superior generalization to fluid manipulation tasks. Thus, DaXBench’s approach achieves state-of-the-art results in both precise control and adaptability across diverse DOM challenges.
Model-free RL, common in deep RL, demands significant data, limiting real-world applications. 

Model-based RL, however, is more data-efficient when a model is available. 


\textbf{Reducing RL Iteration Costs} To decrease the cost of strategy iteration, initializing RL with manual demonstrations is often beneficial.

Chen et al proposes DiffSRL \cite{chen2022diffsrl}, Learning Dynamical State Representation for DOM with Differentiable Simulation through a novel reinforcement learning (RL) approach, achieving state-of-the-art performance by using physically informed state representation learning. DiffSRL incorporates a differentiable physics engine into its pipeline, allowing it to learn physical constraints directly through gradients. This physics-informed learning is crucial for DOM, where maintaining physical properties like non-penetration and continuity is challenging but necessary for realistic manipulation.

DiffSRL uses model-free reinforcement learning (MFRL), specifically the Twin Delayed Deep Deterministic Policy Gradient (TD3) algorithm, which is known for stability in continuous control tasks. With TD3, the policy network is trained on a latent space generated by DiffSRL's encoder, where latent states are informed by differentiable simulation to capture key physical dynamics. This enables the model to produce actions that respect the object’s physical constraints, significantly improving performance in trajectory planning and manipulation.

\textbf{Addressing Simulation-to-Reality Gaps}: Although simulators are valuable for initial training, there is often a mismatch when applying learned strategies to real-world settings. Domain Randomization (DR) addresses this by introducing random visual variations in the simulation to better align with real-world visual differences.

To close the sim2Real gap, DiffSRL \cite{chen2022diffsrl} adapts from simulated to real-world environments by augmenting training data with noise and reducing particle density, thus preparing the model for real sensor data, which is often sparse or noisy. Despite its strengths, limitations remain, such as handling interactions between multiple deformable objects and extensive deformations. Overall, DiffSRL’s gradient-based, physics-aware approach significantly enhances DOM by incorporating physical constraints, accelerating training, and achieving more robust policy learning.

Despite the success of domain randomization (DR) in enabling cross-domain generalization, challenges remain in tailoring strategies to specific tasks and adapting them effectively. \cite{matas2018} addressed this by implementing a visual reinforcement learning (RL) pipeline with pixel-level domain adaptation, effectively bridging simulation and real-world performance in a medical task involving tissue retraction. Another valuable approach is learning a residual strategy to address the simulation-reality gap. For instance, Lv et al. \cite{lv2022} demonstrated successful real-world needle threading by employing a residual strategy, showing that this method can significantly ease the transition from simulation to reality.

\subsubsection{Imitation Learning for DOM}
Imitation Learning (IL) is a control method that learns a policy from expert demonstrations, mapping raw inputs to actions. IL has proven effective in various manipulation tasks. IL aims to derive a policy that accurately replicates expert behavior. This can be expressed mathematically as:

\begin{equation} \label{eq:11}
\pi^*=\underset{\pi \in \Pi}{\arg \min } \mathbb{D}\left(p\left(\pi^{\text {demo }}\right) \| p(\pi)\right)
\end{equation}

where \( D \) represents a divergence metric measuring the difference between two distributions, which could be distributions of state-action pairs, trajectory features, or feature expectations. Unlike RL, IL does not require a reward function, thus avoiding the complexities of reward design, which can be challenging in robotic applications. However, IL requires extensive demonstration data and may struggle with generalizing beyond learned instances.

Early IL research focused on teaching robots primitive motions for DOM through kinesthetic guidance. For instance, \cite{kudoh2015} developed a method for a two-armed robot to learn aerial knot-tying by breaking the task into six fundamental motions—grasping, releasing, double grasping, wrapping, twisting, and sliding learned through human demonstrations. This approach extends to tasks like rope tying , fabric folding , and stitching. Task-parameterized models can further improve generality, adapting actions like dressing motions to various arm positions.

More recent work enables robots to plan actions based on current observations by leveraging large annotated datasets. For example, one approach uses 4,300 labeled human actions to train a model that can predict jaw-clamping poses for folding garments, achieving a 93\% success rate in 120 seconds on average . However, IL requires retraining to transfer skills to new instances. For example, a hat-wearing model must undergo thousands of repetitions in training, both simulated and real, to learn the task. To adapt to a new hat with different shape and flexibility, the model must collect new data and undergo retraining to account for these variations.

As a result, these steps introduce substantial computational and time requirements, limiting the practicality of this approach in real-world applications. Recent research has examined imitation learning (IL) for manipulation at the category level with rigid objects. However, these methods don’t extend to deformable objects (DOs). Unlike rigid objects, whose positions can be represented in low-dimensional space, DOs have a vast range of possible configurations and are prone to self-occlusion, making it challenging to generalize skills across different deformable objects.

To address these challenges, \cite{ren2023} introduced an innovative framework for category-level manipulation of deformable 3D objects, allowing skills learned on one object to transfer to similar new objects with just a single demonstration. This framework includes two key components: the Nocs State Transition (NST) and Neuro-Spatial Encoding (NSE). NST aligns the observed target point cloud into a predefined, standardized pose (the "Nocs state"), which forms a consistent basis for learning skills across a category of objects. NSE then adapts these learned skills to new instances by encoding spatial information at the category level, enabling the system to determine the optimal grasping point for different objects without requiring retraining.


\begin{table}[ht]
\scriptsize 
\centering
\caption{Overview of recent literature in planning and Control for deformable object manipulation}
\label{tab:three}
\begin{tabular}{l p{4cm} p{4cm} p{4cm}}

\hline
\textbf{Planning and Control} & \textbf{Advantages}                                                                                                            & \textbf{Disadvantages}                                                                                                        & \textbf{Literature}                                                                                                                                                                               \\ \hline
\textbf{Analytical Approaches} & 
\begin{itemize}[noitemsep, leftmargin=0.7em]
    \item High precision and predictability.
    \item Robust, systematic frameworks.
    \item Tailored, reproducible solutions.
    \item Scalable to variations.
\end{itemize} & 
\begin{itemize}[noitemsep, leftmargin=0.7em]
    \item Poor adaptability to uncertainties.
    \item High computational cost.
    \item Relies on accurate models.
    \item Complex development.
\end{itemize} & 
\begin{itemize}[noitemsep, leftmargin=0.7em]
    \item \cite{Zhi2024}
    \item \cite{Deng2024}
    \item \cite{Choi2024}
\end{itemize} \\ \hline
\textbf{Reinforcement Learning} & 
\begin{itemize}[noitemsep, leftmargin=0.7em]
    \item Adaptive and robust.
    \item Excels in dynamic scenarios.
    \item Reduces need for explicit models.
    \item Handles complex, high-dimensional tasks.
\end{itemize} & 
\begin{itemize}[noitemsep, leftmargin=0.7em]
    \item High training costs.
    \item Needs large, diverse data.
    \item Struggles with sim-to-real gaps.
    \item Risk of overfitting.
\end{itemize} & 
\begin{itemize}[noitemsep, leftmargin=0.7em]
    \item \cite{Gieselmann2024}
    \item \cite{Yang2024}
    \item \cite{Galassi2024}
\end{itemize} \\ \hline
\textbf{Imitation Learning} & 
\begin{itemize}[noitemsep, leftmargin=0.7em]
    \item Adapts to deformable dynamics.
    \item Requires fewer data.
    \item Task-specific optimization.
    \item Sim2Real benefits.
\end{itemize} & 
\begin{itemize}[noitemsep, leftmargin=0.7em]
    \item Needs high-quality demos.
    \item Limited generalization.
    \item Computationally heavy.
    \item Risk of overfitting.
\end{itemize} & 
\begin{itemize}[noitemsep, leftmargin=0.7em]
    \item \cite{Choi2024}
    \item \cite{Galassi2024}
    \item \cite{Winter2024}
\end{itemize} \\ \hline
\end{tabular}
\end{table}


\subsection{Summary on Manipulation}

The integration of physically informed reinforcement learning (RL) and expert demonstrations has proven effective in addressing critical challenges in deformable object manipulation (DOM). Specifically, physically informed RL enables the precise control necessary for DOM tasks by embedding physical dynamics within the RL framework. For instance, Chen et al. demonstrate in their DaXBench framework that combining RL with differentiable physics significantly enhances control precision. In high-precision tasks, such as the "Whip-Rope" manipulation, advanced RL methods like Analytic Policy Gradients (APG) and Short-Horizon Actor-Critic (SHAC) outperformed traditional methods like Proximal Policy Optimization (PPO), achieving near-optimal control levels due to the model’s physics-informed guidance. This approach showcases the capacity of physically informed RL to handle the intricate, non-linear behaviors inherent in deformable objects.

Additionally, physically informed RL enables orders of magnitude faster convergence and more efficient learning. In DaXBench, the incorporation of physical dynamics into the RL learning process substantially accelerates training and enhances adaptability. For example, in complex manipulation tasks such as "Pour-Water," the SHAC method achieved a significantly higher performance reward compared to PPO, demonstrating the benefits of a physics-aware model in reducing training time and improving task proficiency. This efficiency exemplifies how embedding physical constraints directly into RL models can drive rapid learning, particularly when addressing the complex state-action spaces typical in DOM.

Moreover, the use of expert demonstrations within RL frameworks further accelerates convergence and enhances learning efficiency. In sparse-reward tasks like cloth folding, expert demonstrations provide essential initial guidance, effectively reducing the exploration phase and focusing learning trajectories on productive pathways. The integration of expert trajectories in imitation learning tasks led to notable performance improvements, with tasks like "Fold-T-shirt" reaching high reward levels (e.g., a reward of 0.85 with Imitation Learning via Differentiable Physics, ILD). This illustrates how leveraging expert demonstrations can dramatically expedite convergence by establishing an efficient foundation for learning without requiring extensive trial and error.

Finally, expert demonstrations enable the execution of long-horizon, complex, and multi-stage tasks within RL frameworks. By defining task specifications through expert examples, this approach circumvents the challenges associated with purely exploratory learning, which may lead to efficient yet undesirable or unsafe solutions. Expert-guided trajectories allow for precise task definitions, ensuring that the RL agent avoids hazardous pathways and adheres to safe, feasible task completion strategies. This technique is particularly advantageous in handling long-horizon or multi-stage tasks in DOM, which are otherwise challenging due to the vast and variable state spaces involved. Collectively, these advancements in physically informed RL and expert demonstrations underscore their essential role in advancing the state-of-the-art in efficient and effective deformable object manipulation. For those seeking literature on notable recent advancements in manipulation, please see Table \ref{tab:three}.
\begin{itemize}
    \item Physically informed Reinforcement Learning enables precise control needed for DOM.
    \item Physically informed RL enables orders of magnitude faster convergence and more efficient learning.
    \item Expert demonstrations enable orders of magnitude faster convergence and more efficient learning.
    \item Expert demonstrations enable long-horizon or complex multi-stage tasks.
\end{itemize}

\section{Discussion}

Referring back to the challenges outlined at the beginning of this review and comparing them with the identified gaps from the summaries, we propose the following directions for future research. These are structured according to the key challenges (see Section~\ref{key-challenges}).

\begin{figure*}[ht] 
    \centering
    \includegraphics[width=\textwidth]{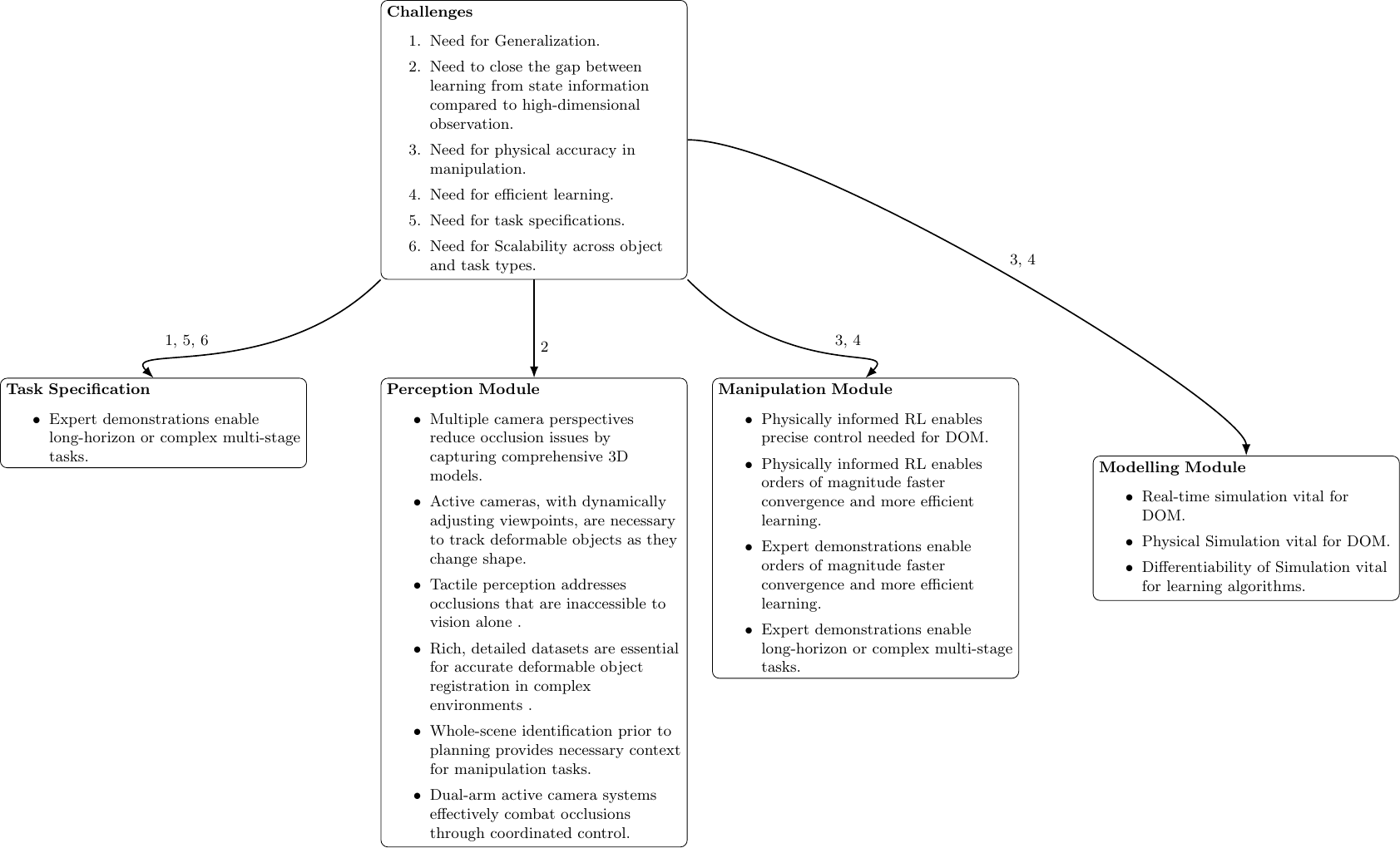}
    \caption{Illustration links techniques and approaches that show great promise in DOM to face the key challenges outlined at the beginning of this review.}
    \label{fig:proposal}
\end{figure*}


By combining multiple camera perspectives with an active camera and tactile perception into one perception system there is great potential to reduce occlusions (the main source of noise in perception data) in unstructured environments to a level we have not yet seen in DOM. In addition, perception systems have been shown to benefit greatly from performing whole scene identification prior to attempting a task, particularly for long-horizon or complex multi-stage tasks. For the purpose of scene identification and active tracking, the creation of rich simulated datasets with deform-ability annotations would greatly benefit registering deformable objects and their associated physical qualities (elasticity, plasticity, etc) and quantities (Young's Modulus, Poisson Ratio, etc). A system such as this provides maximal data to the manipulation and modeling parts of a DOM framework (see Perception Module in Figure \ref{fig:proposal} ). Enabling extensive domain randomization upon simulated sensor data could lead to significantly more robust policies in real experiments and applications. Recent innovations have made a previously computationally intensive technique suitable for real-time application, NERF \cite{Mildenhall2020} could be applied to perception in DOM since it excels at capturing whole scenes in high detail. This includes the capacity for novel view synthesis and the use of multiple camera perspectives. In addition NERF only requires an RGB camera, which doesn't suffer from the drawbacks of RGBD such as the dampening effect upon the depth sensor when interacting with soft-objects; a key drawback to using RBGD in DOM. The highly detailed representations achievable with NERF can also collect textural details of objects, which may allow for better classification of material properties even before leveraging tactile data. NERF also enables the automatic generation of highly detailed 3D models, simplifying the task of simulated data collection.


Evidently physically-informed reinforcement learning, that is, Reinforcement Learning agents trained within differentiable simulations show orders of magnitude improved performance on several metrics, including; learning efficiency, precision in low-level control and generalization. Further research should explore leveraging expert demonstrations to accelerate convergence and to mitigate the unproductive exploration that typically occurs when RL agents are trained without such guided seed trajectories.

A broader innovation could be creating expert demonstrations within simulation, drastically reducing labor and equipment costs associated with doing this work in the real world. By using the simulation as a data collection tool we may be able to make use of expert demonstrations as a form of task specification. Defining not just the goal states, which are notoriously difficult to describe for deformable objects, but also the manner in which the task should be carried out. A large data set of this sort, covering a broad range of tasks and objects could have immense value to the field of DOM. Combining this approach with recent advancements in generative networks \cite{Goodfellow2014} and large language models \cite{Vaswani2023} these task specifying simulated expert demonstrations could perhaps be generated as easily as large language models now generate human-level text. Advancements in generative networks have demonstrated effectiveness across text, images, 3-D models \cite{Shi2023} according to the survey by Shi et al. Therefore it is reasonable to believe the application of generative networks could extend to deformable objects and digital twins more generally. This would open many doors, enabling a huge diversity of tasks and deformable object scenarios to be trained quickly and entirely automated. Simulated expert demonstrations as task specifications could become a kind of standardized data format for Reinforcement Learning approaches that are specialized to work with this sort of data. Zargarbashi et al. recently did something similar, albeit outside DOM, but within Robotics simulation \cite{Zargarbashi2024}. A further innovation that is well supported is the use of graph neural networks for high-level decision making, the branching characteristic of these networks directly mimics algorithmic behavior of conventional computer programming languages \cite{Xu2019} and even the networks of the human brain. A possible approach could be to use the sequences of simulation states of the simulated expert demonstrations as nodes in a graph neural network for high-level planning in DOM. This could make it possible to map and index key steps in the manipulators interactions and trajectory. This may simplify failure detection and make more straightforward the task of initiating re-grasping and other forms of dynamic or spontaneous decision making in response to trajectories that vary too widely from the next expected state. Furthermore, many of the core elements in deformable object manipulation, including scene representations, simulation trajectories, mesh structures, and deformability annotations can be naturally expressed as graph-structured data. This makes graph neural networks a compelling unifying framework for integrating heterogeneous information across tasks. Such representations may enable models that generalize more effectively across diverse deformable object scenarios, rather than being tailored to narrowly defined task instances.

\section{Conclusion}

The field of deformable object manipulation (DOM) has made significant strides, yet several challenges remain, particularly in addressing occlusions, enabling generalization across tasks and object types, and achieving scalable, real-time solutions. This review highlights promising advancements and proposes future directions rooted in the integration of perception, modeling, and manipulation.

The combination of multi-camera perspectives, active vision systems, and tactile perception shows potential for significantly reducing occlusions in unstructured environments. Enhancements such as whole-scene identification and rich simulated datasets with detailed deformability annotations can further support robust and adaptive perception systems. The adoption of neural radiance fields (NERFs) may prove transformative in implementing such a perception system, offering high-detail scene capture with reduced hardware demands.

Physically informed reinforcement learning (RL) has emerged as a critical tool, with differentiable simulations improving efficiency, precision, and generalization. Incorporating expert demonstrations into these frameworks can accelerate convergence and address the complex task specifications unique to DOM. Simulated expert demonstrations offer a scalable approach to defining task specifications, particularly when combined with generative neural networks which could be able generalize to unseen object-task combinations given enough scale. In addition, task specifications (sequences of simulation states) have great potential as a standardized data format for RL applications in DOM. Finally, leveraging advanced computational frameworks such as graph neural networks for high-level decision-making and integrating simulated datasets for training can help bridge the gap between simulation and real-world application. By addressing these challenges, DOM research can unlock new possibilities across industries, creating adaptable robotic systems capable of handling diverse and dynamic deformable objects.

\section{Funding sources}
This work was supported by a Departmental Scholarship under the EPSRC DigiCORTEX Project, \textit{Deep Learning of Deformable Object Manipulation on Manufacturing Robots} [EP/W014688/1].

The funding source had no role in the study design; collection, analysis, or interpretation of data; writing of the manuscript; or the decision to submit the article for publication.

\bibliographystyle{elsarticle-num}
\bibliography{myrefs}

@ARTICLE{kroemer2021review,
  author  = {Kroemer, O. and Niekum, S. and Konidaris, G.},
  title   = {A review of robot learning for manipulation: Challenges, representations, and algorithms},
  journal = {The Journal of Machine Learning Research},
  volume  = {22},
  number  = {1},
  pages   = {1395--1476},
  year    = {2021}
}

@ARTICLE{zhu2022challenges,
  author  = {Zhu, J. and others},
  title   = {Challenges and outlook in robotic manipulation of deformable objects},
  journal = {IEEE Robotics and Automation Magazine},
  volume  = {29},
  number  = {3},
  pages   = {67--77},
  year    = {2022},
  note    = {doi: 10.1109/MRA.2022.3147415}
}

@ARTICLE{gao2022hierarchical,
  author  = {Gao, Y. and Chen, Z. and Ling, Y. and Yang, J. and Liu, Y.-H. and Li, X.},
  title   = {A hierarchical manipulation scheme for robotic sorting of multiwire cables with hybrid vision},
  journal = {IEEE/ASME Transactions on Mechatronics},
  year    = {2022}
}

@ARTICLE{li2018vision,
  author  = {Li, X. and Su, X. and Liu, Y.-H.},
  title   = {Vision-based robotic manipulation of flexible pcbs},
  journal = {IEEE/ASME Transactions on Mechatronics},
  volume  = {23},
  number  = {6},
  pages   = {2739--2749},
  year    = {2018}
}

@ARTICLE{jin2021trajectory,
  author  = {Jin, S. and Romeres, D. and Ragunathan, A. and Jha, D. K. and Tomizuka, M.},
  title   = {Trajectory optimization for manipulation of deformable objects: Assembly of belt drive units},
  journal = {2021 IEEE International Conference on Robotics and Automation (ICRA)},
  pages   = {10002--10008},
  year    = {2021},
  note    = {IEEE}
}

@ARTICLE{zhou2020practical,
  author  = {Zhou, H. and Li, S. and Lu, Q. and Qian, J.},
  title   = {A practical solution to deformable linear object manipulation: A case study on cable harness connection},
  journal = {2020 5th International Conference on Advanced Robotics and Mechatronics (ICARM)},
  pages   = {329--333},
  year    = {2020},
  note    = {IEEE}
}

@ARTICLE{shi2023robocook,
  author  = {Shi, H. and Xu, H. and Clarke, S. and Li, Y. and Wu, J.},
  title   = {Robocook: Long-horizon elasto-plastic object manipulation with diverse tools},
  journal = {arXiv preprint},
  year    = {2023},
  note    = {arXiv:2306.14447}
}

@ARTICLE{xu2023roboninja,
  author  = {Xu, Z. and Xian, Z. and Lin, X. and Chi, C. and Huang, Z. and Gan, C. and Song, S.},
  title   = {Roboninja: Learning an adaptive cutting policy for multi-material objects},
  journal = {arXiv preprint},
  year    = {2023},
  note    = {arXiv:2302.11553}
}

@ARTICLE{narasimhan2022self,
  author  = {Narasimhan, G. and Zhang, K. and Eisner, B. and Lin, X. and Held, D.},
  title   = {Self-supervised transparent liquid segmentation for robotic pouring},
  journal = {2022 International Conference on Robotics and Automation (ICRA)},
  pages   = {4555--4561},
  year    = {2022},
  note    = {IEEE}
}

@ARTICLE{billard2019trends,
  author  = {Billard, A. and Kragic, D.},
  title   = {Trends and challenges in robot manipulation},
  journal = {Science},
  volume  = {364},
  number  = {6446},
  pages   = {eaat8414},
  year    = {2019}
}

@ARTICLE{yin2021modeling,
  author  = {Yin, H. and Varava, A. and Kragic, D.},
  title   = {Modeling, learning, perception, and control methods for deformable object manipulation},
  journal = {Science Robotics},
  volume  = {6},
  number  = {54},
  pages   = {eabd8803},
  year    = {2021}
}

@ARTICLE{kadi2023data,
  author  = {Kadi, H. A. and Terzić, K.},
  title   = {Data-driven robotic manipulation of cloth-like deformable objects: The present, challenges and future prospects},
  journal = {Sensors},
  volume  = {23},
  number  = {5},
  pages   = {2389},
  year    = {2023}
}

@ARTICLE{lv2020review,
  author  = {Lv, N. and Liu, J. and Xia, H. and Ma, J. and Yang, X.},
  title   = {A review of techniques for modeling flexible cables},
  journal = {Computer-Aided Design},
  volume  = {122},
  pages   = {102826},
  year    = {2020}
}

@ARTICLE{hou2019review,
  author  = {Hou, Y. C. and Sahari, K. S. M. and How, D. N. T.},
  title   = {A review on modeling of flexible deformable object for dexterous robotic manipulation},
  journal = {International Journal of Advanced Robotic Systems},
  volume  = {16},
  number  = {3},
  pages   = {1729881419848894},
  year    = {2019}
}

@ARTICLE{arriola2020modeling,
  author  = {Arriola-Rios, V. E. and Guler, P. and Ficuciello, F. and Kragic, D. and Siciliano, B. and Wyatt, J. L.},
  title   = {Modeling of deformable objects for robotic manipulation: A tutorial and review},
  journal = {Frontiers in Robotics and AI},
  volume  = {7},
  pages   = {82},
  year    = {2020}
}

@ARTICLE{sanchez2018robotic,
  author  = {Sanchez, J. and Corrales, J.-A. and Bouzgarrou, B.-C. and Mezouar, Y.},
  title   = {Robotic manipulation and sensing of deformable objects in domestic and industrial applications: a survey},
  journal = {The International Journal of Robotics Research},
  volume  = {37},
  number  = {7},
  pages   = {688--716},
  year    = {2018}
}

@ARTICLE{gu_survey,
  author  = {Gu, F. and Zhou, Y. and Wang, Z. and Jiang, S. and He, B.},
  title   = {A survey on robotic manipulation of deformable objects: Recent advances, open challenges and new frontiers},
  journal = {Unpublished or preprint},
  year    = {n.d.}
}

@ARTICLE{luo2018deep,
  author  = {Luo, J. and Solowjow, E. and Wen, C. and Ojea, J. A. and Agogino, A. M.},
  title   = {Deep reinforcement learning for robotic assembly of mixed deformable and rigid objects},
  journal = {2018 IEEE/RSJ International Conference on Intelligent Robots and Systems (IROS)},
  pages   = {2062--2069},
  year    = {2018},
  note    = {IEEE}
}

@ARTICLE{sun2015accurate,
  author  = {Sun, L. and Aragon-Camarasa, G. and Rogers, S. and Siebert, J. P.},
  title   = {Accurate garment surface analysis using an active stereo robot head with application to dual-arm flattening},
  journal = {2015 IEEE International Conference on Robotics and Automation (ICRA)},
  pages   = {185--192},
  year    = {2015},
  note    = {IEEE}
}

@ARTICLE{lin2023softgym,
  author  = {Lin, X. and Wang, Y. and Olkin, J. and Held, D.},
  title   = {SoftGym: Benchmarking deep reinforcement learning for deformable object manipulation},
  journal = {arXiv preprint},
  year    = {2023},
  note    = {Accessed: Oct. 09, 2023. Available: https://arxiv.org/abs/2011.07215v2}
}

@ARTICLE{mataslearning,
  author  = {Matas, J. and Davidson, A. and Johns, E. and James, S.},
  title   = {Learning end-to-end robotic manipulation of deformable objects},
  journal = {Unpublished or preprint},
  year    = {n.d.}
}

@ARTICLE{liang2024real,
  author  = {Liang, X. and Liu, F. and Zhang, Y. and Li, Y. and Lin, S. and Yip, M.},
  title   = {Real-to-sim deformable object manipulation: Optimizing physics models with residual mappings for robotic surgery},
  journal = {arXiv preprint},
  year    = {2024},
  note    = {arXiv:2309.11656. Accessed: Oct. 14, 2024. Available: http://arxiv.org/abs/2309.11656}
}

@ARTICLE{valencia2019toward,
  author  = {Valencia, A. J. and Nadon, F. and Payeur, P.},
  title   = {Toward real-time 3D shape tracking of deformable objects for robotic manipulation and shape control},
  journal = {2019 IEEE SENSORS},
  pages   = {1--4},
  year    = {2019},
  note    = {Montreal, QC, Canada: IEEE. doi: 10.1109/SENSORS43011.2019.8956623}
}

@ARTICLE{lee2020,
  author  = {Lee, M. A. and Zhu, Y. and Zachares, P. and Tan, M. and Srinivasan, K. and Savarese, S. and Fei-Fei, L. and Garg, A. and Bohg, J.},
  title   = {Making sense of vision and touch: Learning multimodal representations for contact-rich tasks},
  journal = {IEEE Transactions on Robotics},
  volume  = {36},
  number  = {3},
  pages   = {582--596},
  year    = {2020}
}

@ARTICLE{battaglia2016,
  author  = {Battaglia, P. and Pascanu, R. and Lai, M. and Jimenez Rezende, D. and others},
  title   = {Interaction networks for learning about objects, relations and physics},
  journal = {Advances in Neural Information Processing Systems},
  volume  = {29},
  year    = {2016}
}

@ARTICLE{nadon2020,
  author  = {Nadon, F. and Valencia, A. J. and Sambandam, N. and Rowlands, S. and Dickens, J. and Payeur, P.},
  title   = {Toward a General Framework for 3D Deformable Object Grasping and Manipulation},
  journal = {Workshop on Robotic Manipulation of Deformable Objects (ROMADO) at the IEEE/RSJ International Conference on Intelligent Robots and Systems (IROS)},
  year    = {2020}
}

@ARTICLE{henrich2023,
  author  = {Henrich, P. and Gyenes, B. and Scheikl, P. M. and Neumann, G. and Mathis-Ullrich, F.},
  title   = {Registered and Segmented Deformable Object Reconstruction from a Single View Point Cloud},
  journal = {Proc. 2024 IEEE/CVF Winter Conference on Applications of Computer Vision (WACV)},
  year    = {2023},
  note    = {arXiv:2311.07357. Accessed: Nov. 04, 2024. Available: http://arxiv.org/abs/2311.07357}
}

@ARTICLE{chen2024,
  author  = {Chen, S. and Xu, Y. and Yu, C. and Li, L. and Hsu, D.},
  title   = {Differentiable Particles for General-Purpose Deformable Object Manipulation},
  journal = {arXiv},
  year    = {2024},
  note    = {arXiv:2405.01044. Accessed: Nov. 04, 2024. Available: http://arxiv.org/abs/2405.01044}
}

@ARTICLE{weng2024,
  author  = {Weng, Z. and others},
  title   = {Interactive Perception for Deformable Object Manipulation},
  journal = {arXiv},
  year    = {2024},
  note    = {arXiv:2403.05177. Accessed: Nov. 04, 2024. Available: http://arxiv.org/abs/2403.05177}
}

@ARTICLE{newcombe2015,
  author  = {Newcombe, R. A. and Fox, D. and Seitz, S. M.},
  title   = {DynamicFusion: Reconstruction and tracking of non-rigid scenes in real-time},
  journal = {Proceedings of the IEEE Conference on Computer Vision and Pattern Recognition},
  pages   = {343--352},
  year    = {2015}
}

@ARTICLE{dou2017,
  author  = {Dou, M. and Davidson, P. and Fanello, S. R. and Khamis, S. and Kowdle, A. and Rhemann, C. and Tankovich, V. and Izadi, S.},
  title   = {Motion2Fusion: Real-time volumetric performance capture},
  journal = {ACM Transactions on Graphics (ToG)},
  volume  = {36},
  number  = {6},
  pages   = {1--16},
  year    = {2017}
}

@ARTICLE{collins2016,
  author  = {Collins, T. and Bartoli, A. and Bourdel, N. and Canis, M.},
  title   = {Robust, real-time, dense and deformable 3D organ tracking in laparoscopic videos},
  journal = {International Conference on Medical Image Computing and Computer-Assisted Intervention},
  pages   = {404--412},
  year    = {2016},
  note    = {Springer}
}

@ARTICLE{haouchine2013,
  author  = {Haouchine, N. and Dequidt, J. and Peterlik, I. and Kerrien, E. and Berger, M.-O. and Cotin, S.},
  title   = {Image-guided simulation of heterogeneous tissue deformation for augmented reality during hepatic surgery},
  journal = {2013 IEEE International Symposium on Mixed and Augmented Reality (ISMAR)},
  pages   = {199--208},
  year    = {2013},
  note    = {IEEE}
}

@ARTICLE{schulman2013,
  author  = {Schulman, J. and Lee, A. and Ho, J. and Abbeel, P.},
  title   = {Tracking deformable objects with point clouds},
  journal = {2013 IEEE International Conference on Robotics and Automation},
  pages   = {1130--1137},
  year    = {2013},
  note    = {IEEE}
}

@ARTICLE{petit2017Pizza,
  author  = {Petit, A. and Lippiello, V. and Fontanelli, G. A. and Siciliano, B.},
  title   = {Tracking elastic deformable objects with an RGB-D sensor for a pizza chef robot},
  journal = {Robotics and Autonomous Systems},
  volume  = {88},
  pages   = {187--201},
  year    = {2017}
}

@ARTICLE{tang2017,
  author  = {Tang, T. and Fan, Y. and Lin, H.-C. and Tomizuka, M.},
  title   = {State estimation for deformable objects by point registration and dynamic simulation},
  journal = {2017 IEEE/RSJ International Conference on Intelligent Robots and Systems (IROS)},
  pages   = {2427--2433},
  year    = {2017},
  note    = {IEEE}
}

@ARTICLE{tang2018,
  author  = {Tang, T. and Wang, C. and Tomizuka, M.},
  title   = {A framework for manipulating deformable linear objects by coherent point drift},
  journal = {IEEE Robotics and Automation Letters},
  volume  = {3},
  number  = {4},
  pages   = {3426--3433},
  year    = {2018}
}

@ARTICLE{myronenko2010,
  author  = {Myronenko, A. and Song, X.},
  title   = {Point set registration: Coherent point drift},
  journal = {IEEE Transactions on Pattern Analysis and Machine Intelligence},
  volume  = {32},
  number  = {12},
  pages   = {2262--2275},
  year    = {2010}
}

@ARTICLE{narang2021,
  author  = {Narang, Y. S. and Sundaralingam, B. and Van Wyk, K. and Mousavian, A. and Fox, D.},
  title   = {Interpreting and predicting tactile signals for the SynTouch BioTac},
  journal = {The International Journal of Robotics Research},
  volume  = {40},
  number  = {12-14},
  pages   = {1467--1487},
  year    = {2021}
}

@ARTICLE{sundaram2019,
  author  = {Sundaram, S. and Kellnhofer, P. and Li, Y. and Zhu, J.-Y. and Torralba, A. and Matusik, W.},
  title   = {Learning the signatures of the human grasp using a scalable tactile glove},
  journal = {Nature},
  volume  = {569},
  number  = {7758},
  pages   = {698--702},
  year    = {2019}
}

@ARTICLE{yuan2017,
  author  = {Yuan, W. and Dong, S. and Adelson, E. H.},
  title   = {GelSight: High-resolution robot tactile sensors for estimating geometry and force},
  journal = {Sensors},
  volume  = {17},
  number  = {12},
  pages   = {2762},
  year    = {2017}
}

@ARTICLE{she2021,
  author  = {She, Y. and Wang, S. and Dong, S. and Sunil, N. and Rodriguez, A. and Adelson, E.},
  title   = {Cable manipulation with a tactile-reactive gripper},
  journal = {The International Journal of Robotics Research},
  volume  = {40},
  number  = {12-14},
  pages   = {1385--1401},
  year    = {2021}
}

@ARTICLE{hellman2017,
  author  = {Hellman, R. B. and Tekin, C. and van der Schaar, M. and Santos, V. J.},
  title   = {Functional contour-following via haptic perception and reinforcement learning},
  journal = {IEEE Transactions on Haptics},
  volume  = {11},
  number  = {1},
  pages   = {61--72},
  year    = {2017}
}

@ARTICLE{zheng2022,
  author  = {Zheng, Y. and Veiga, F. F. and Peters, J. and Santos, V. J.},
  title   = {Autonomous learning of page flipping movements via tactile feedback},
  journal = {IEEE Transactions on Robotics},
  volume  = {38},
  number  = {5},
  pages   = {2734--2749},
  year    = {2022}
}

@ARTICLE{zhang2021,
  author  = {Zhang, Q. and Li, Y. and Luo, Y. and Shou, W. and Foshey, M. and Yan, J. and Tenenbaum, J. B. and Matusik, W. and Torralba, A.},
  title   = {Dynamic modeling of hand-object interactions via tactile sensing},
  journal = {2021 IEEE/RSJ International Conference on Intelligent Robots and Systems (IROS)},
  pages   = {2874--2881},
  year    = {2021},
  note    = {IEEE}
}

@ARTICLE{zhou2023enhancing,
  author  = {Zhou, P.},
  title   = {Enhancing Deformable Object Manipulation By Using Interactive Perception and Assistive Tools},
  journal = {arXiv},
  year    = {2023},
  month   = {Nov. 16},
  note    = {arXiv:2311.09659},
  doi     = {10.48550/arXiv.2311.09659}
}

@ARTICLE{zhang2023interaction,
  author  = {Zhang, H. and Lu, Z. and Liang, W. and Yu, H. and Mao, Y. and Wu, Y.},
  title   = {Interaction Control for Tool Manipulation on Deformable Objects Using Tactile Feedback},
  journal = {IEEE Robotics and Automation Letters},
  volume  = {PP},
  pages   = {1--8},
  year    = {2023},
  month   = {May},
  doi     = {10.1109/LRA.2023.3257680}
}

@ARTICLE{yu2023precise,
  author  = {Yu, Z. and others},
  title   = {Precise Robotic Needle-Threading with Tactile Perception and Reinforcement Learning},
  journal = {Unpublished},
  year    = {2023},
  note    = {Details not provided}
}

@ARTICLE{xie2023hmdo,
  author  = {Xie, W. and Yu, Z. and Zhao, Z. and Zuo, B. and Wang, Y.},
  title   = {HMDO: Markerless Multi-view Hand Manipulation Capture with Deformable Objects},
  journal = {arXiv},
  year    = {2023},
  month   = {Jan. 18},
  note    = {arXiv:2301.07652},
  doi     = {10.48550/arXiv.2301.07652}
}

@ARTICLE{wu2023learning,
  author  = {Wu, R. and Ning, C. and Dong, H.},
  title   = {Learning Foresightful Dense Visual Affordance for Deformable Object Manipulation},
  journal = {arXiv},
  year    = {2023},
  month   = {Jul. 21},
  note    = {arXiv:2303.11057},
  doi     = {10.48550/arXiv.2303.11057}
}

@ARTICLE{wi2024virdo++,
  author  = {Wi, Y. and Zeng, A. and Florence, P. and Fazeli, N.},
  title   = {VIRDO++: Real-World, Visuo-Tactile Dynamics and Perception of Deformable Objects},
  journal = {Unpublished},
  year    = {2024},
  note    = {Details not provided}
}

@ARTICLE{thach2024deformernet,
  author  = {Thach, B. and Cho, B. Y. and Ho, S.-H. and Hermans, T. and Kuntz, A.},
  title   = {DeformerNet: Learning Bimanual Manipulation of 3D Deformable Objects},
  journal = {arXiv},
  year    = {2024},
  month   = {Feb. 19},
  note    = {arXiv:2305.04449},
  doi     = {10.48550/arXiv.2305.04449}
}

@ARTICLE{deng2023learning,
  author  = {Deng, Y. and Wang, X. and Chen, L.},
  title   = {Learning visual-based deformable object rearrangement with local graph neural networks},
  journal = {Complex Intelligent Systems},
  volume  = {9},
  number  = {5},
  pages   = {5923--5936},
  year    = {2023},
  month   = {Oct.},
  doi     = {10.1007/s40747-023-01048-w}
}

@ARTICLE{muller2008,
  author  = {Müller, M. and Stam, J. and James, D. and Thürey, N.},
  title   = {Real time physics: Class notes},
  journal = {ACM SIGGRAPH 2008 Classes, SIGGRAPH},
  year    = {2008},
  pages   = {88:1--88:90},
  publisher = {ACM}
}

@ARTICLE{bender2017,
  author  = {Bender, J. and Müller, M. and Macklin, M.},
  title   = {Position-based simulation methods in computer graphics},
  journal = {EUROGRAPHICS Tutorials},
  year    = {2017},
  editor  = {Zwicker, M. and Soler, C.},
  publisher = {Eurographics Association}
}

@INCOLLECTION{faure2012sofa,
  author    = {Faure, F. and Duriez, C. and Delingette, H. and Allard, J. and Gilles, B. and Marchesseau, S. and Talbot, H. and Courtecuisse, H. and Bousquet, G. and Peterlik, I. and others},
  title     = {SOFA: A Multi-Model Framework for Interactive Physical Simulation},
  booktitle = {Soft Tissue Biomechanical Modeling for Computer Assisted Surgery},
  editor    = {Payan, Y.},
  publisher = {Springer},
  address   = {Berlin, Heidelberg},
  pages     = {283--321},
  year      = {2012},
  note      = {doi: 10.1007/8415\_2012\_125}
}

@MISC{sofa2023performance,
  author  = {{SOFA Consortium}},
  title   = {Improving Performance in SOFA Simulations},
  howpublished = {\url{https://sofa-framework.github.io/doc/using-sofa/performances/improve-performances/}},
  year    = {2023},
  note    = {Accessed: 2025-10-07}
}

@INPROCEEDINGS{duriez2018softrobots,
  author    = {Duriez, C. and Coevoet, E. and Largilliere, F. and Dequidt, J. and Marchal, M. and Kruszewski, A. and Goury, O.},
  title     = {SOFA and Soft Robotics: The SOFA SoftRobot Plugin},
  booktitle = {IEEE International Conference on Soft Robotics (RoboSoft)},
  pages     = {711--716},
  year      = {2018},
  note      = {doi: 10.1109/ROBOSOFT.2018.8404955}
}

@ARTICLE{nunes2024ppo,
  author  = {Nunes, R. and Oliveira, A. and Costa, P. and Lima, P.},
  title   = {Development of a PPO-Reinforcement Learned Walking Tripedal Soft-Legged Robot Using SOFA},
  journal = {arXiv preprint arXiv:2504.09242},
  year    = {2024},
  note    = {Available at: \url{https://arxiv.org/abs/2504.09242}}
}

@ARTICLE{makoviychuk2021,
  author  = {Makoviychuk, V. and others},
  title   = {Isaac Gym: High Performance GPU Based Physics Simulation For Robot Learning},
  journal = {Thirty-fifth Conference on Neural Information Processing Systems Datasets and Benchmarks Track (Round 2)},
  year    = {2021},
  month   = {Aug.},
  note    = {Accessed: Oct. 21, 2023},
  url     = {https://openreview.net/forum?id=fgFBtYgJQX\_}
}

@ARTICLE{huang2021,
  author  = {Huang, I. and others},
  title   = {DefGraspSim: Simulation-based grasping of 3D deformable objects},
  journal = {arXiv},
  year    = {2021},
  month   = {Jul. 12},
  note    = {arXiv:2107.05778, Accessed: Nov. 04, 2024},
  url     = {http://arxiv.org/abs/2107.05778}
}

@ARTICLE{schulman2013case,
  author  = {Schulman, J. and Gupta, A. and Venkatesan, S. and Tayson-Frederick, M. and Abbeel, P.},
  title   = {A case study of trajectory transfer through non-rigid registration for a simplified suturing scenario},
  journal = {Proceedings of the IEEE/RSJ International Conference on Intelligent Robots and Systems (IROS)},
  year    = {2013},
  pages   = {4111--4117},
  publisher = {IEEE}
}

@ARTICLE{kita2011,
  author  = {Kita, Y. and Kanehiro, F. and Ueshiba, T. and Kita, N.},
  title   = {Clothes handling based on recognition by strategic observation},
  journal = {Proceedings of the IEEE-RAS International Conference on Humanoid Robots},
  year    = {2011},
  pages   = {53--58},
  publisher = {IEEE}
}

@ARTICLE{macklin2014,
  author  = {Macklin, M. and Müller, M. and Chentanez, N. and Kim, T.-Y.},
  title   = {Unified particle physics for real-time applications},
  journal = {ACM Transactions on Graphics},
  volume  = {33},
  pages   = {153},
  year    = {2014}
}

@ARTICLE{macklin2016,
  author  = {Macklin, M. and Müller, M. and Chentanez, N.},
  title   = {XPBD: Position-based simulation of compliant constrained dynamics},
  journal = {Proceedings of the 9th International Conference on Motion in Games, MIG '16},
  year    = {2016},
  pages   = {49--54},
  publisher = {ACM}
}

@ARTICLE{guler2015,
  author  = {Güler, P. and Pauwels, K. and Pieropan, A. and Kjellström, H. and Kragic, D.},
  title   = {Estimating the deformability of elastic materials using optical flow and position-based dynamics},
  journal = {Proceedings of the IEEE International Conference on Humanoid Robots (Humanoids)},
  year    = {2015},
  pages   = {965--971},
  publisher = {IEEE}
}

@ARTICLE{boonvisut2013,
  author  = {Boonvisut, P. and Çavuşoğlu, M. C.},
  title   = {Estimation of soft tissue mechanical parameters from robotic manipulation data},
  journal = {IEEE/ASME Transactions on Mechatronics},
  volume  = {18},
  pages   = {1602--1611},
  year    = {2013}
}

@ARTICLE{zollhofer2014,
  author  = {Zollhöfer, M. and Nießner, M. and Izadi, S. and Rhemann, C. and Zach, C. and Fisher, M. and Wu, C. and Fitzgibbon, A. and Loop, C. and Theobalt, C. and Stamminger, M.},
  title   = {Real-time non-rigid reconstruction using an RGB-D camera},
  journal = {ACM Transactions on Graphics},
  volume  = {33},
  pages   = {156},
  year    = {2014}
}

@ARTICLE{faure2012,
  author  = {Faure, F. and Duriez, C. and Delingette, H. and Allard, J. and Gilles, B. and Marchesseau, S. and Talbot, H. and Courtecuisse, H. and Bousquet, G. and Peterlik, I. and Cotin, S.},
  title   = {SOFA: A multi-model framework for interactive physical simulation},
  journal = {Soft Tissue Biomechanical Modeling for Computer Assisted Surgery},
  volume  = {11},
  pages   = {283--321},
  year    = {2012},
  editor  = {Payan, Y.},
  publisher = {Springer},
  series  = {Studies in Mechanobiology, Tissue Engineering and Biomaterials}
}

@ARTICLE{yoshida2015,
  author  = {Yoshida, E. and Ayusawa, K. and Ramirez-Alpizar, I. G. and Harada, K. and Duriez, C. and Kheddar, A.},
  title   = {Simulation-based optimal motion planning for deformable object},
  journal = {IEEE International Workshop on Advanced Robotics and its Social Impacts (ARSO)},
  year    = {2015},
  pages   = {1--6},
  publisher = {IEEE}
}

@ARTICLE{matas2018,
  author  = {Matas, J. and James, S. and Davison, A. J.},
  title   = {Sim-to-real reinforcement learning for deformable object manipulation},
  journal = {Proceedings of the Conference on Robot Learning (CoRL)},
  pages   = {734--743},
  year    = {2018},
  publisher = {PMLR}
}

@ARTICLE{mcconachie2018,
  author  = {McConachie, D. and Berenson, D.},
  title   = {Estimating model utility for deformable object manipulation using multiarmed bandit methods},
  journal = {IEEE Transactions on Automation Science and Engineering},
  volume  = {15},
  pages   = {967--979},
  year    = {2018}
}

@ARTICLE{petit2017,
  author  = {Petit, A. and Ficuciello, F. and Fontanelli, G. A. and Villani, L. and Siciliano, B.},
  title   = {Using physical modeling and RGB-D registration for contact force sensing on deformable objects},
  journal = {International Conference on Informatics in Control, Automation and Robotics (ICINCO)},
  volume  = {2},
  pages   = {24--33},
  year    = {2017},
  publisher = {Springer}
}

@ARTICLE{haouchine2018,
  author  = {Haouchine, N. and Kuang, W. and Cotin, S. and Yip, M.},
  title   = {Vision-based force feedback estimation for robot-assisted surgery using instrument-constrained biomechanical three-dimensional maps},
  journal = {IEEE Robotics and Automation Letters},
  volume  = {3},
  pages   = {2160--2165},
  year    = {2018}
}

@ARTICLE{nvidia2019,
  author  = {{NVIDIA}},
  title   = {PhysX SDK},
  journal = {Online},
  year    = {2019},
  note    = {Available: https://developer.nvidia.com/physx-sdk. Accessed: 25-Sep-2019}
}

@ARTICLE{bai2016,
  author  = {Bai, Y. and Yu, W. and Liu, C. K.},
  title   = {Dexterous manipulation of cloth},
  journal = {Proceedings of the 37th Annual Conference of the European Association for Computer Graphics, EG '16},
  pages   = {523--532},
  year    = {2016},
  publisher = {Eurographics Association}
}

@ARTICLE{yu2017,
  author  = {Yu, W. and Kapusta, A. and Tan, J. and Kemp, C. C. and Turk, G. and Liu, C. K.},
  title   = {Haptic simulation for robot-assisted dressing},
  journal = {Proceedings of the IEEE International Conference on Robotics and Automation (ICRA)},
  pages   = {6044--6051},
  year    = {2017},
  publisher = {IEEE}
}

@ARTICLE{erickson2018,
  author  = {Erickson, Z. and Clever, H. M. and Turk, G. and Liu, C. K. and Kemp, C. C.},
  title   = {Deep haptic model predictive control for robot-assisted dressing},
  journal = {2018 IEEE International Conference on Robotics and Automation (ICRA)},
  pages   = {4437--4444},
  year    = {2018},
  publisher = {IEEE}
}

@ARTICLE{kapusta2019,
  author  = {Kapusta, A. and Erickson, Z. and Clever, H. M. and Yu, W. and Liu, C. K. and Turk, G. and Kemp, C. C.},
  title   = {Personalized collaborative plans for robot-assisted dressing via optimization and simulation},
  journal = {Autonomous Robots},
  volume  = {43},
  pages   = {2183--2207},
  year    = {2019}
}

@ARTICLE{mordatch2012,
  author  = {Mordatch, I. and Todorov, E. and Popović, Z.},
  title   = {Discovery of complex behaviors through contact-invariant optimization},
  journal = {ACM Transactions on Graphics},
  volume  = {31},
  pages   = {43},
  year    = {2012}
}

@ARTICLE{brockman2016,
  author  = {Brockman, G. and Cheung, V. and Pettersson, L. and Schneider, J. and Schulman, J. and Tang, J. and Zaremba, W.},
  title   = {OpenAI Gym},
  journal = {arXiv preprint arXiv:1606.01540},
  year    = {2016}
}

@ARTICLE{todorov2014,
  author  = {Todorov, E.},
  title   = {Convex and analytically-invertible dynamics with contacts and constraints: Theory and implementation in MuJoCo},
  journal = {2014 IEEE International Conference on Robotics and Automation (ICRA)},
  pages   = {6054--6061},
  year    = {2014},
  publisher = {IEEE}
}

@ARTICLE{petrik2019,
  author  = {Petrík, V. and Kyrki, V.},
  title   = {Feedback-based fabric strip folding},
  journal = {2019 IEEE/RSJ International Conference on Intelligent Robots and Systems (IROS)},
  pages   = {773--778},
  year    = {2019},
  publisher = {IEEE}
}

@ARTICLE{yan2020,
  author  = {Yan, W. and Vangipuram, A. and Abbeel, P. and Pinto, L.},
  title   = {Learning predictive representations for deformable objects using contrastive estimation},
  journal = {Proceedings of the Conference on Robot Learning (CoRL)},
  year    = {2020},
  publisher = {PMLR}
}

@ARTICLE{coumans2016,
  author  = {Coumans, E. and Bai, Y.},
  title   = {PyBullet, a Python module for physics simulation for games, robotics and machine learning},
  journal = {Online},
  year    = {2016},
  note    = {Available: http://pybullet.org}
}

@ARTICLE{elbrechter2012,
  author  = {Elbrechter, C. and Haschke, R. and Ritter, H.},
  title   = {Folding paper with anthropomorphic robot hands using real-time physics-based modeling},
  journal = {Proceedings of IEEE International Conference on Humanoid Robots (Humanoids)},
  pages   = {210--215},
  year    = {2012},
  publisher = {IEEE}
}

@ARTICLE{erickson2020,
  author  = {Erickson, Z. and Gangaram, V. and Kapusta, A. and Liu, C. K. and Kemp, C. C.},
  title   = {Assistive Gym: A physics simulation framework for assistive robotics},
  journal = {Proceedings of the IEEE International Conference on Robotics and Automation (ICRA)},
  pages   = {10169--10176},
  year    = {2020},
  publisher = {IEEE}
}

@ARTICLE{Zaidi2020,
  author  = {Zaidi, L. and Corrales Ramon, J. A. and Sabourin, L. and Bouzgarrou, B. C. and Mezouar, Y.},
  title   = {Grasp Planning Pipeline for Robust Manipulation of 3D Deformable Objects with Industrial Robotic Hand + Arm Systems},
  journal = {Applied Sciences},
  volume  = {10},
  number  = {23},
  year    = {2020},
  pages   = {Article 23},
  doi     = {10.3390/app10238736}
}

@ARTICLE{Yu2023,
  author  = {Yu, M. and Lv, K. and Wang, C. and Tomizuka, M. and Li, X.},
  title   = {A Coarse-to-Fine Framework for Dual-Arm Manipulation of Deformable Linear Objects with Whole-Body Obstacle Avoidance},
  journal = {2023 IEEE International Conference on Robotics and Automation (ICRA)},
  year    = {2023},
  pages   = {10153--10159},
  doi     = {10.1109/ICRA48891.2023.10160264}
}

@ARTICLE{Tabata2023,
  author  = {Tabata, K. and Seki, H. and Tsuji, T. and Hiramitsu, T.},
  title   = {Mass spring model for non-uniformed deformable linear object toward dexterous manipulation},
  journal = {Artificial Life and Robotics},
  volume  = {28},
  number  = {4},
  pages   = {812--822},
  year    = {2023},
  doi     = {10.1007/s10015-023-00889-5}
}

@ARTICLE{Shi2024,
  author  = {Shi, D. and Hu, H. and Yang, C. and Lu, Z. and Li, Q.},
  title   = {A Learning System for Deformable Object Cooperative Manipulation},
  journal = {IEEE Transactions on Automation Science and Engineering},
  pages   = {1--12},
  year    = {2024},
  doi     = {10.1109/TASE.2024.3486063}
}

@ARTICLE{Sanchez2020,
  author  = {Sanchez, J. and Mohy El Dine, K. and Corrales, J. A. and Bouzgarrou, B.-C. and Mezouar, Y.},
  title   = {Blind Manipulation of Deformable Objects Based on Force Sensing and Finite Element Modeling},
  journal = {Frontiers in Robotics and AI},
  volume  = {7},
  year    = {2020},
  doi     = {10.3389/frobt.2020.00073}
}

@ARTICLE{Liu2023,
  author  = {Liu, F. and Su, E. and Lu, J. and Li, M. and Yip, M. C.},
  title   = {Robotic Manipulation of Deformable Rope-Like Objects Using Differentiable Compliant Position-Based Dynamics},
  journal = {IEEE Robotics and Automation Letters},
  volume  = {8},
  number  = {7},
  pages   = {3964--3971},
  year    = {2023},
  doi     = {10.1109/LRA.2023.3264766}
}

@ARTICLE{chen2022daxbench,
  author  = {Chen, S. and others},
  title   = {DaxBench: Benchmarking Deformable Object Manipulation with Differentiable Physics},
  journal = {International Conference on Learning Representations (ICLR)},
  year    = {2022},
  note    = {Accessed: Jan. 17, 2024},
  url     = {https://openreview.net/forum?id=1NAzMofMnWl}
}

@ARTICLE{chen2022diffsrl,
  author  = {Chen, S. and Liu, Y. and Li, J. and Yao, S. W. and Fan, T. and Pan, J.},
  title   = {DiffSRL: Learning Dynamical State Representation for Deformable Object Manipulation with Differentiable Simulator},
  journal = {arXiv preprint arXiv:2110.12352},
  year    = {2022},
  month   = {Jul.},
  doi     = {10.48550/arXiv.2110.12352}
}

@ARTICLE{huang2023,
  author  = {Huang, Y. and Xia, C. and Wang, X. and Liang, B.},
  title   = {Learning graph dynamics with external contact for deformable linear objects shape control},
  journal = {IEEE Robotics and Automation Letters},
  year    = {2023}
}

@ARTICLE{shi2022,
  author  = {Shi, H. and Xu, H. and Huang, Z. and Li, Y. and Wu, J.},
  title   = {RoboCraft: Learning to see, simulate, and shape elasto-plastic objects with graph networks},
  journal = {arXiv preprint arXiv:2205.02909},
  year    = {2022}
}

@ARTICLE{zaidi2017,
  author  = {Zaidi, L. and Corrales, J. A. and Bouzgarrou, B. C. and Mezouar, Y. and Sabourin, L.},
  title   = {Model-based strategy for grasping 3D deformable objects using a multi-fingered robotic hand},
  journal = {Robotics and Autonomous Systems},
  volume  = {95},
  pages   = {196--206},
  year    = {2017}
}

@ARTICLE{li2015,
  author  = {Li, Y. and Yue, Y. and Xu, D. and Grinspun, E. and Allen, P. K.},
  title   = {Folding deformable objects using predictive simulation and trajectory optimization},
  journal = {2015 IEEE/RSJ International Conference on Intelligent Robots and Systems (IROS)},
  year    = {2015},
  pages   = {6000--6006}
}

@ARTICLE{lin2015,
  author  = {Lin, H. and Guo, F. and Wang, F. and Jia, Y.-B.},
  title   = {Picking up a soft 3D object by 'feeling' the grip},
  journal = {The International Journal of Robotics Research},
  volume  = {34},
  number  = {11},
  pages   = {1361--1384},
  year    = {2015}
}

@ARTICLE{sintov2020,
  author  = {Sintov, A. and Macenski, S. and Borum, A. and Bretl, T.},
  title   = {Motion planning for dual-arm manipulation of elastic rods},
  journal = {IEEE Robotics and Automation Letters},
  volume  = {5},
  number  = {4},
  pages   = {6065--6072},
  year    = {2020}
}

@ARTICLE{lui2013,
  author  = {Lui, W. H. and Saxena, A.},
  title   = {Tangled: Learning to untangle ropes with RGB-D perception},
  journal = {2013 IEEE/RSJ International Conference on Intelligent Robots and Systems (IROS)},
  year    = {2013}
}

@ARTICLE{mcconachie2020,
  author  = {McConachie, D. and Power, T. and Mitrano, P. and Berenson, D.},
  title   = {Learning when to trust a dynamics model for planning in reduced state spaces},
  journal = {IEEE Robotics and Automation Letters},
  volume  = {5},
  number  = {2},
  pages   = {3540--3547},
  year    = {2020}
}

@ARTICLE{bretl2014,
  author  = {Bretl, T. and McCarthy, Z.},
  title   = {Quasi-static manipulation of a Kirchhoff elastic rod based on a geometric analysis of equilibrium configurations},
  journal = {The International Journal of Robotics Research},
  volume  = {33},
  number  = {1},
  pages   = {48--68},
  year    = {2014}
}

@ARTICLE{shah2016,
  author  = {Shah, A. J. and Shah, J. A.},
  title   = {Towards manipulation planning for multiple interlinked deformable linear objects},
  journal = {2016 IEEE International Conference on Robotics and Automation (ICRA)},
  publisher = {IEEE},
  pages   = {3908--3915},
  year    = {2016}
}

@ARTICLE{lv2022,
  author  = {Lv, J. and Feng, Y. and Zhang, C. and Zhao, S. and Shao, L. and Lu, C.},
  title   = {SAMRL: Sensing-aware model-based reinforcement learning via differentiable physics-based simulation and rendering},
  journal = {arXiv preprint arXiv:2210.15185},
  year    = {2022}
}

@ARTICLE{kudoh2015,
  author  = {Kudoh, S. and Gomi, T. and Katano, R. and Tomizawa, T. and Suehiro, T.},
  title   = {In-air knotting of rope by a dual-arm multi-finger robot},
  journal = {2015 IEEE/RSJ International Conference on Intelligent Robots and Systems (IROS)},
  publisher = {IEEE},
  pages   = {6202--6207},
  year    = {2015}
}

@ARTICLE{ren2023,
  author  = {Ren, Y. and Chen, R. and Cong, Y.},
  title   = {Autonomous manipulation learning for similar deformable objects via only one demonstration},
  journal = {Proceedings of the IEEE/CVF Conference on Computer Vision and Pattern Recognition},
  pages   = {17069--17078},
  year    = {2023}
}

@article{Zhi2024,
  author  = {H. Zhi and others},
  title   = {Non-Prehensile Object Transport by Nonholonomic Robots Connected by Linear Deformable Elements},
  journal = {IEEE Robotics and Automation Letters},
  volume  = {9},
  number  = {10},
  pages   = {8651--8658},
  year    = {2024},
  month   = {Oct.},
  doi     = {10.1109/LRA.2024.3440096}
}

@article{Yang2024,
  author  = {Z. Yang and Y. Wang and Y. Jiang and H. Zhang and C. Yang},
  title   = {DeformerNet based 3D Deformable Objects Shape Servo Control for Bimanual Robot Manipulation},
  journal = {2024 IEEE International Conference on Industrial Technology (ICIT)},
  pages   = {1--7},
  year    = {2024},
  month   = {Mar.},
  doi     = {10.1109/ICIT58233.2024.10570080}
}

@article{Winter2024,
  author  = {T. R. Winter and A. M. Sundaram and W. Friedl and M. A. Roa and F. Stulp and J. Silvério},
  title   = {State- and context-dependent robotic manipulation and grasping via uncertainty-aware imitation learning},
  journal = {arXiv},
  year    = {2024},
  month   = {Oct.},
  note    = {arXiv:2410.24035},
  doi     = {10.48550/arXiv.2410.24035}
}

@article{Gieselmann2024,
  author  = {R. Gieselmann},
  title   = {Synergies between Policy Learning and Sampling-based Planning},
  journal = {Technical Report},
  year    = {2024},
  note    = {Accessed: Nov. 24, 2024},
  url     = {https://urn.kb.se/resolve?urn=urn:nbn:se:kth:diva-341911}
}

@article{Galassi2024,
  author  = {K. Galassi and B. Wu and J. Perez and G. Palli and J.-M. Renders},
  title   = {Attention-Based Cloth Manipulation from Model-free Topological Representation},
  journal = {2024 IEEE International Conference on Robotics and Automation (ICRA)},
  pages   = {18207--18213},
  year    = {2024},
  month   = {May},
  doi     = {10.1109/ICRA57147.2024.10610241}
}

@article{Deng2024,
  author  = {H. Deng and F. Ahmad and J. Xiong and Z. Xia},
  title   = {A Robot-Object Unified Modeling Method for Deformable Object Manipulation in Constrained Environments},
  journal = {IEEE/ASME Transactions on Mechatronics},
  pages   = {1--12},
  year    = {2024},
  doi     = {10.1109/TMECH.2024.3371111}
}

@article{Choi2024,
  author  = {A. Choi and D. Tong and D. Terzopoulos and J. Joo and M. K. Jawed},
  title   = {Learning Neural Force Manifolds for Sim2Real Robotic Symmetrical Paper Folding},
  journal = {IEEE Transactions on Automation Science and Engineering},
  pages   = {1--0},
  year    = {2024},
  doi     = {10.1109/TASE.2024.3366909}
}

@article{shiach2024,
  author  = {J. Shiach},
  title   = {Shooting method for ODEs},
  journal = {Online resource},
  year    = {2024},
  note    = {Accessed: Nov. 24, 2024},
  url     = {https://jonshiach.github.io/ODEs-book/_pages/5.1_Shooting_method.html}
}

@article{tedrake_underactuated,
  author  = {R. Tedrake},
  title   = {Underactuated Robotics: Algorithms for Walking, Running, Swimming, Flying, and Manipulation},
  journal = {MIT Course Notes for 6.832},
  year    = {2023},
  note    = {Accessed: Nov. 24, 2024},
  url     = {https://underactuated.csail.mit.edu/trajectory_optimization.html}
}

@article{spong2005,
  author  = {M. W. Spong and S. Hutchinson and M. Vidyasagar},
  title   = {Robot Modeling and Control},
  journal = {Wiley},
  volume  = {1st ed.},
  address = {Hoboken, NJ, USA},
  year    = {2005}
}

@article{sutton2018,
  author  = {R. S. Sutton and A. G. Barto},
  title   = {Reinforcement Learning: An Introduction},
  journal = {MIT Press},
  volume  = {2nd ed.},
  address = {Cambridge, MA, USA},
  year    = {2018}
}

@article{Goodfellow2014,
  author  = {I. J. Goodfellow and others},
  title   = {Generative adversarial networks},
  journal = {arXiv},
  year    = {2014},
  month   = {Jun.},
  note    = {arXiv:1406.2661},
  doi     = {10.48550/arXiv.1406.2661}
}

@article{Vaswani2023,
  author  = {A. Vaswani and others},
  title   = {Attention is all you need},
  journal = {arXiv},
  year    = {2023},
  month   = {Aug.},
  note    = {arXiv:1706.03762},
  doi     = {10.48550/arXiv.1706.03762}
}

@article{Zargarbashi2024,
  author  = {F. Zargarbashi and J. Cheng and D. Kang and R. Sumner and S. Coros},
  title   = {RobotKeyframing: Learning locomotion with high-level objectives via mixture of dense and sparse rewards},
  journal = {arXiv},
  year    = {2024},
  month   = {Nov.},
  note    = {arXiv:2407.11562, Accessed: Nov. 12, 2024},
  url     = {http://arxiv.org/abs/2407.11562}
}

@article{Mildenhall2020,
  author  = {B. Mildenhall and P. P. Srinivasan and M. Tancik and J. T. Barron and R. Ramamoorthi and R. Ng},
  title   = {NeRF: Representing scenes as neural radiance fields for view synthesis},
  journal = {arXiv},
  year    = {2020},
  month   = {Aug.},
  note    = {arXiv:2003.08934, Accessed: Oct. 14, 2024},
  url     = {http://arxiv.org/abs/2003.08934}
}

@article{Xu2019,
  author  = {K. Xu and W. Hu and J. Leskovec and S. Jegelka},
  title   = {How powerful are graph neural networks?},
  journal = {arXiv},
  year    = {2019},
  month   = {Feb.},
  note    = {arXiv:1810.00826},
  doi     = {10.48550/arXiv.1810.00826}
}

@article{Shi2023,
  author  = {Z. Shi and S. Peng and Y. Xu and A. Geiger and Y. Liao and Y. Shen},
  title   = {Deep generative models on 3D representations: A survey},
  journal = {arXiv},
  year    = {2023},
  month   = {Aug.},
  note    = {arXiv:2210.15663, Accessed: Oct. 14, 2024},
  url     = {http://arxiv.org/abs/2210.15663}
}

\end{document}